\documentclass[10pt,twocolumn,letterpaper]{article}

\usepackage{authblk}

\usepackage[pagenumbers]{cvpr} 

\usepackage[dvipsnames]{xcolor}

\newcommand{\citeMain}{\cite}
\newcommand{\citeSupp}{\cite}
\usepackage{bm}
\usepackage[T1]{fontenc} 
\usepackage[accsupp]{axessibility} 
\definecolor{cvprblue}{rgb}{0.21,0.49,0.74}
\usepackage[pagebackref,breaklinks,colorlinks,citecolor=cvprblue]{hyperref}

\def\paperID{1329} 
\def\confName{CVPR}
\def\confYear{2024}

\title{SleepVST: Sleep Staging from Near-Infrared Video Signals\\using Pre-Trained Transformers}

\author[1,2]{Jonathan F. Carter\thanks{\texttt{jcarter@robots.ox.ac.uk}}}
\author[2]{Jo\~{a}o Jorge}
\author[2]{Oliver Gibson}
\author[1]{Lionel Tarassenko}
\affil[1]{Institute of Biomedical Engineering, University of Oxford}
\affil[2]{Oxehealth Ltd., Oxford}

\makeatletter
\newcommand{\phantomlabel}[2]{
    \protected@write\@auxout{}{
        \string\newlabel{#2}{
            {\@currentlabel#1}{\thepage}
            {\@currentlabel#1}{#2}{}
        }
    }
    \hypertarget{#2}{}
}
\makeatother

\begin{document}
\maketitle
\begin{abstract}
    Advances in camera-based physiological monitoring have enabled the robust, non-contact measurement of respiration and the cardiac pulse, which are known to be indicative of the sleep stage. This has led to research into camera-based sleep monitoring as a promising alternative to ``gold-standard'' polysomnography, which is cumbersome, expensive to administer, and hence unsuitable for longer-term clinical studies. In this paper, we introduce SleepVST, a transformer model which enables state-of-the-art performance in camera-based sleep stage classification (sleep staging). After pre-training on contact sensor data, SleepVST outperforms existing methods for cardio-respiratory sleep staging on the SHHS and MESA datasets, achieving total Cohen's kappa scores of 0.75 and 0.77 respectively. We then show that SleepVST can be successfully transferred to cardio-respiratory waveforms extracted from video, enabling fully contact-free sleep staging. Using a video dataset of 50 nights, we achieve a total accuracy of 78.8\% and a Cohen's $\kappa$ of 0.71 in four-class video-based sleep staging, setting a new state-of-the-art in the domain.
\end{abstract}
\section{Introduction}\label{section:intro}
Accurate sleep monitoring is critical for the diagnosis of sleep disorders and the discovery of novel treatments and biomarkers. Poor sleep is broadly linked with a number of health conditions, including cardiovascular diseases such as diabetes and hypertension~\citeMain{hafner_why_2017}. Additionally, there are links between specific sleep stages and neurodegenerative conditions, such as a decrease in non-rapid-eye-movement stage 3 (N3) sleep with Alzheimer's disease~\citeMain{lee_slow_2020}.

Overnight video polysomnography (vPSG), the “gold-standard”~\citeMain{stefani_prospective_2015} technique for sleep monitoring, requires the use of a large number of contact sensors including electrodes placed on the scalp (EEG), near the eyes (EOG), and under the chin (EMG) of the patient. Human experts (polysomnographers) must then review the recorded data, classifying the patient's sleep into five stages at 30-second intervals (epochs) and annotating other important events such as leg movements and apnoeas. This manual process often takes multiple hours to complete.

Experts primarily rely on characteristic patterns in brain activity measured from the EEG to classify sleep stages~\citeMain{iber_aasm_2007}. However, sleep stage information is also encoded in measures of autonomic activity, including cardiac~\citeMain{shinar_automatic_2001} and respiratory signals~\citeMain{hudgel_mechanics_1984}, and body movements~\citeMain{wilde-frenz_rate_1983}. This has led to the development of methods for automatically classifying sleep stages from sensors that measure these parameters, such as smartwatches~\citeMain{walch_sleep_2019}.

Prior work has shown that these physiological parameters can also be measured using video cameras~\citeMain{wang_algorithmic_2017, wang_algorithmic_2022, heinrich_body_2013}, leading to the development of methods for classifying sleep stages entirely from video input~\citeMain{nochino_sleep_2019, van_meulen_contactless_2023, carter_deep_2023}. Camera-based methods have particular promise as part of a multi-purpose sleep monitoring system. For example, within elderly care settings, in addition to classifying sleep stages, they could also be used to detect specific sleep movement disorders~\citeMain{heinrich_body_2013}, which are linked with conditions such as Parkinson's, and which are typically distinguished via manual video review~\citeMain{stefani_sleep_2020}. They could also be used to detect falls~\citeMain{de_miguel_home_2017}, one of the leading causes of injury and mortality amongst elderly individuals.
\begin{figure*}[t]
  \centering
   \includegraphics[width=1\linewidth]{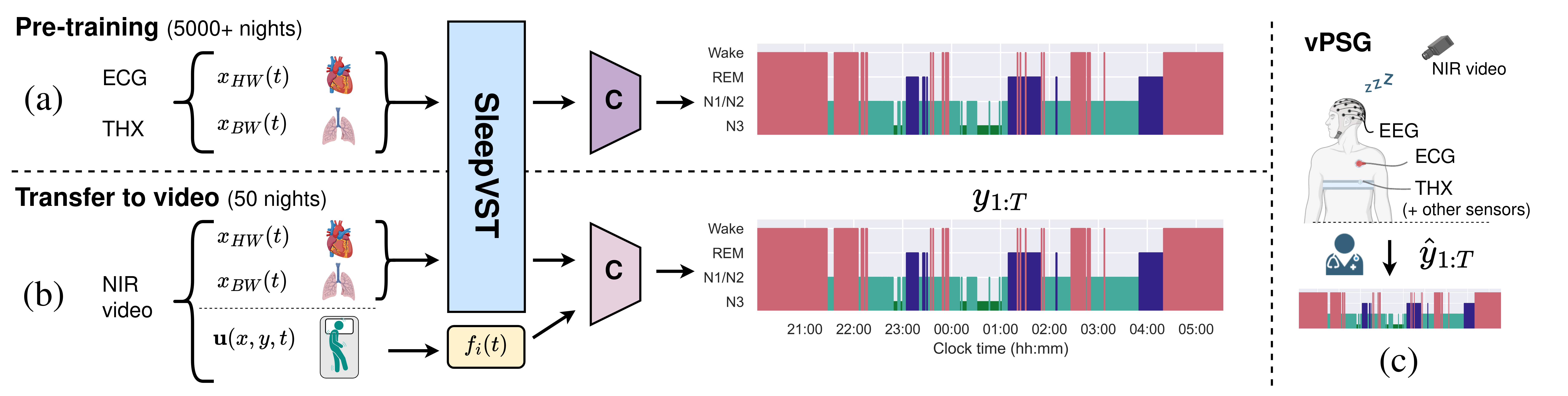}
   \caption{\textbf{Sleep staging from near-infrared video signals using SleepVST.} (a) We first pre-train the model end-to-end on cardiac (heart) and respiratory (breathing) waveforms $x_{HW}(t)$ and $x_{BW}(t)$ derived from the electrocardiogram (ECG) and a thoracic respiratory belt (THX) respectively. (b) After pre-training, we use the model as a frozen feature extractor, applying it to cardio-respiratory waveforms derived from near-infrared (NIR) video to generate sequences of features. When transferring to video data, we additionally use a set of motion features $f_i(t)$ derived from an optical flow field $\bm{u}(x,y,t)$ as inputs to the classifier \textbf{C}. This approach allows us to utilise much larger contact-sensor datasets to train SleepVST, whilst also enabling the incorporation of motion information when transferring to video data. (c) We evaluate SleepVST using overnight video polysomnography (vPSG) studies, comparing expert-labelled sleep stage sequences $\hat{y}_{1:T}$ against those generated entirely from near-infrared video $y_{1:T}$ using our method.}
   \label{fig:transformer}
   \phantomlabel{a}{fig:transformer:a}
   \phantomlabel{b}{fig:transformer:b}
   \phantomlabel{c}{fig:transformer:c}
\end{figure*}

State-of-the-art video-based sleep staging~\citeMain{carter_deep_2023} has used the heart rate (HR) and breathing rate (BR) derived from video to perform sleep staging. However, these \textit{rates} are typically measured as averages over time, losing higher frequency inter-pulse and inter-breath interval information. EEG-~\citeMain{phan_xsleepnet_2022} and wearable-based~\citeMain{kotzen_sleepppg-net_2023} sleep staging methods have both been improved by using raw waveforms as inputs, rather than derived quantities e.g. signal features~\citeMain{pardey_new_1996, redmond_cardiorespiratory-based_2006}.

In this paper, we introduce SleepVST\footnote{Sleep Video-Signal Transformer.}, a transformer model for sleep stage classification from cardio-respiratory waveforms. First, we pre-train the model on waveforms derived from contact sensors (Fig.~\ref{fig:transformer:a}), after which it outperforms existing methods for cardio-respiratory sleep staging on two widely used public datasets, highlighting the effectiveness of our transformer architecture.

After pre-training, we then show that SleepVST can be transferred to cardiac and respiratory waveforms measured entirely from near-infrared video, where our method additionally enables us to incorporate motion features derived from the camera data (Fig.~\ref{fig:transformer:b}). Using this novel combination of inputs available from a near-infrared camera, we obtain state-of-the-art results for video-based sleep staging. Through a series of ablation studies we demonstrate the effectiveness of our approach. Our work further highlights the potential of camera-based sleep monitoring for important clinical applications.
\section{Background and Related Work}

\textbf{Automated sleep staging.}
To reduce the manual effort of sleep assessment, deep learning methods have been developed for the automatic classification of sleep stages from polysomnography (PSG) recordings, achieving near-human accuracy~\citeMain{phan_automatic_2022}. Accuracy is commonly measured using Cohen's kappa statistic ($\kappa$,~\citeMain{cohen_coefficient_1960}), which measures inter-rater agreement between two sources. Expert annotation of sleep stages is a subjective process, making 100\% agreement unachievable~\citeMain{van_gorp_certainty_2022}. Prior work has shown that sleep scorers agree around 80\% of the time ($\kappa$~=~0.71-0.81,~\citeMain{danker-hopfe_interrater_2009,lee_interrater_2022}) when classifying sleep stages according to American Academy of Sleep Medicine (AASM,~\citeMain{iber_aasm_2007}) scoring rules. These rules divide sleep into five classes: Wake, N1 (light), N2 (intermediate), N3 (deep) and rapid-eye-movement (REM) sleep. However, due to low inter-scorer agreement in identifying N1~\citeMain{danker-hopfe_interrater_2009}, prior work has often treated sleep classification as a four-class problem, by merging N1 and N2 into a single class~\citeMain{sridhar_deep_2020, kotzen_sleepppg-net_2023}.

Given the drawbacks of conventional polysomnography, prior work has investigated the potential of automatic sleep staging from a reduced number of contact sensors, including the electrocardiogram (\citeMain{sridhar_deep_2020},~$\kappa$~=~0.66), and the photoplethysmogram (\citeMain{radha_deep_2021},~$\kappa$~=~0.65).

\textbf{Camera-based physiological measurement.}
Respiratory measurement is possible from video by measuring the induced movements of the chest and abdomen~\citeMain{wang_algorithmic_2022}. This has been achieved through methods such as frame-differencing~\citeMain{bai_design_2010, tan_real-time_2010}, using dense optical flow magnitudes~\citeMain{nakajima_method_1997, nakajima_development_2001} and by tracking the movement of sparse optical flow vectors~\citeMain{li_non-contact_2014} across video frames.

Camera-based measurement of the pulse is possible using methods which measure specific cardiac phenomena~\citeMain{mcduff_camera_2023}. These include remote photoplethysmography (rPPG,~\citeMain{poh_advancements_2011, wang_algorithmic_2017}), which measures skin colour changes caused by variations in blood volume over the cardiac cycle, and ballistocardiography~\citeMain{balakrishnan_detecting_2013}, which measures head movements induced as blood is pumped through the carotid arteries. More recently, deep learning methods have been developed which learn to extract a pulse signal directly from input video frames~\citeMain{chen_deepphys_2018, yu_physformer_2022, liu_efficientphys_2023} rather than targeting specific underlying cardiac phenomena, including unsupervised approaches~\citeMain{yang_simper_2023, speth_non-contrastive_2023}.

Camera-based vital-sign measurement has been validated within clinical settings~\citeMain{trumpp_camera-based_2018,villarroel_non-contact_2019}, during overnight polysomnography recordings~\citeMain{van_gastel_camera-based_2021, wang_feasibility_2023}, and has led to products with regulatory clearance for camera-based measurement of the heart and breathing rate, including the Gili Pro Biosensor~\citeMain{us_food_and_drug_administration_fda_novo_2020-1} and Oxehealth Vital Signs software~\citeMain{us_food_and_drug_administration_fda_novo_2020}.

\textbf{Camera-based sleep monitoring.}
Prior work has investigated the binary problem of distinguishing sleep from wake using video cameras, using measures of activity derived from the input frames~\citeMain{schwichtenberg_pediatric_2018, cuppens_automatic_2010, nakajima_development_2001}. Whilst activity information alone can be effective in distinguishing sleep from wake, cardio-respiratory signals carry important additional information for classifying specific sleep stages~\citeMain{hudgel_mechanics_1984, shinar_automatic_2001}. The current state-of-the-art approach for video-based sleep staging (~\citeMain{carter_deep_2023},~$\kappa=0.61$) combines multiple modalities derived from video: the heart rate, breathing rate and activity information, to perform sleep stage classification. This approach outperforms prior methods which had used only a single video-derived modality: motion~\citeMain{nochino_sleep_2019, kamon_development_2022} or the cardiac pulse~\citeMain{van_meulen_contactless_2023}.

\section{Datasets and Preprocessing}
\label{section:datasets}
\subsection{Contact Sensor Data}
\label{section:contactsensors}
To train the SleepVST model, we used the SHHS~\citeMain{quan_sleep_1997, zhang_national_2018} and MESA~\citeMain{chen_racialethnic_2015} contact sensor datasets. These datasets contain PSG recordings and sleep stages annotated according to AASM scoring rules. The cardio-respiratory signals available in each recording include the electrocardiogram (ECG), which measures electrical activity of the heart, and respiratory signals, measured using respiratory bands placed around the abdomen and thorax.

From the available signals, we seek to produce generic cardiac and respiratory waveforms to train the SleepVST model, preventing the model from learning to use information which may be present in signals derived from contact sensors but not present in those derived from video. This is to mitigate the problem of domain shift~\citeMain{shimodaira_improving_2000, luo_taking_2019}, where the performance of a model is significantly reduced when applied to a different data distribution.

To produce a cardiac (heart) waveform, we first applied the same filtering steps used by the Pan-Tompkins algorithm~\citeMain{pan_real-time_1985} to the ECG signal. We resampled the output from the moving-window integration step of the algorithm to 10 Hz, then applied a Butterworth bandpass filter with a passband of 0.66--2.8 Hz (40--168 beats per minute or BPM) and Gaussian smoothing. To produce a respiratory (breathing) waveform, we used the thoracic respiratory signal as the input source, as previously used in~\citeMain{bakker_estimating_2021}. This was downsampled to 5 Hz and smoothed using a median filter with a kernel of length 5, to remove artefacts.

\subsection{Video Data}
\label{section:videodata}
We used the dataset previously introduced in~\citeMain{carter_deep_2023} for our video-based sleep staging experiments. This dataset consists of overnight vPSG recordings from 50 volunteers, with sleep stages annotated according to the AASM guidelines at 30-second intervals (epochs) by a sleep physiologist. Throughout this work, we refer to this as the Oxford Sleep Volunteers (OSV) dataset. Each recording includes 850 nm near-infrared (NIR) video data captured at 20 FPS. An example camera view can be seen in \Cref{fig:flowregions}. More information on the dataset is provided in \S\ref{section:appendix_data}.

To derive cardiac and respiratory waveforms from the video data, we used components of FDA-cleared Oxehealth Vital Signs software~\citeMain{us_food_and_drug_administration_fda_novo_2020}. For each overnight video recording, we used the software to obtain cardiac and respiratory waveforms sampled at 10 Hz and 5 Hz respectively. We then applied the same filtering steps applied to the contact sensor waveforms: Butterworth filtering and Gaussian smoothing to the heart waveforms, and median filtering to the breathing waveforms.

\begin{figure}[htb]
  \centering
   \includegraphics[width=1\linewidth]{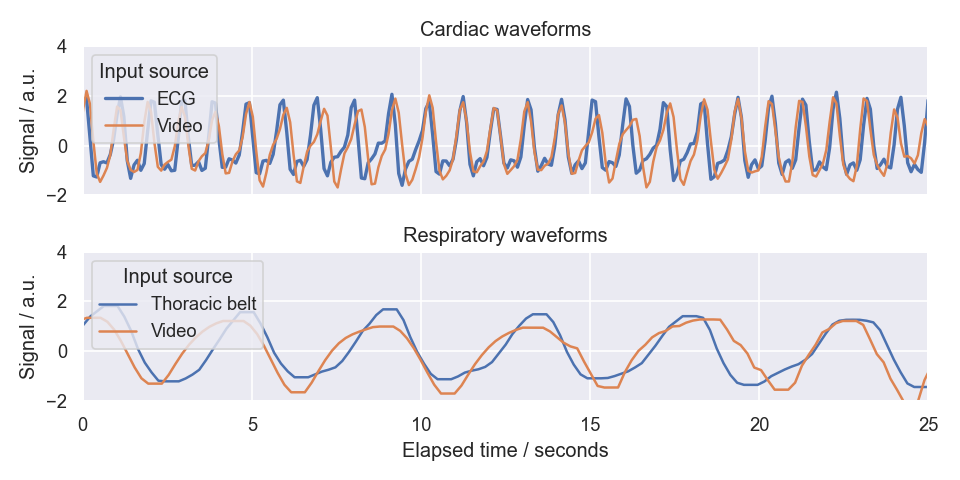}
   \caption{Example (normalized) cardiac and respiratory waveforms, $x_{HW}(t)$ and $x_{BW}(t)$, derived from contact sensors (blue) and video (orange) from the OSV dataset.}\label{fig:signals}
\end{figure}

\Cref{fig:signals} shows examples of processed cardio-respiratory waveforms measured from contact sensors and video for a short section of a recording from the OSV dataset. These have additionally been normalised to zero mean and unit variance.

\section{Methods}
\subsection{SleepVST Architecture}\label{section:vst}
\begin{figure}[thb]
  \centering
   \includegraphics[width=1\linewidth]{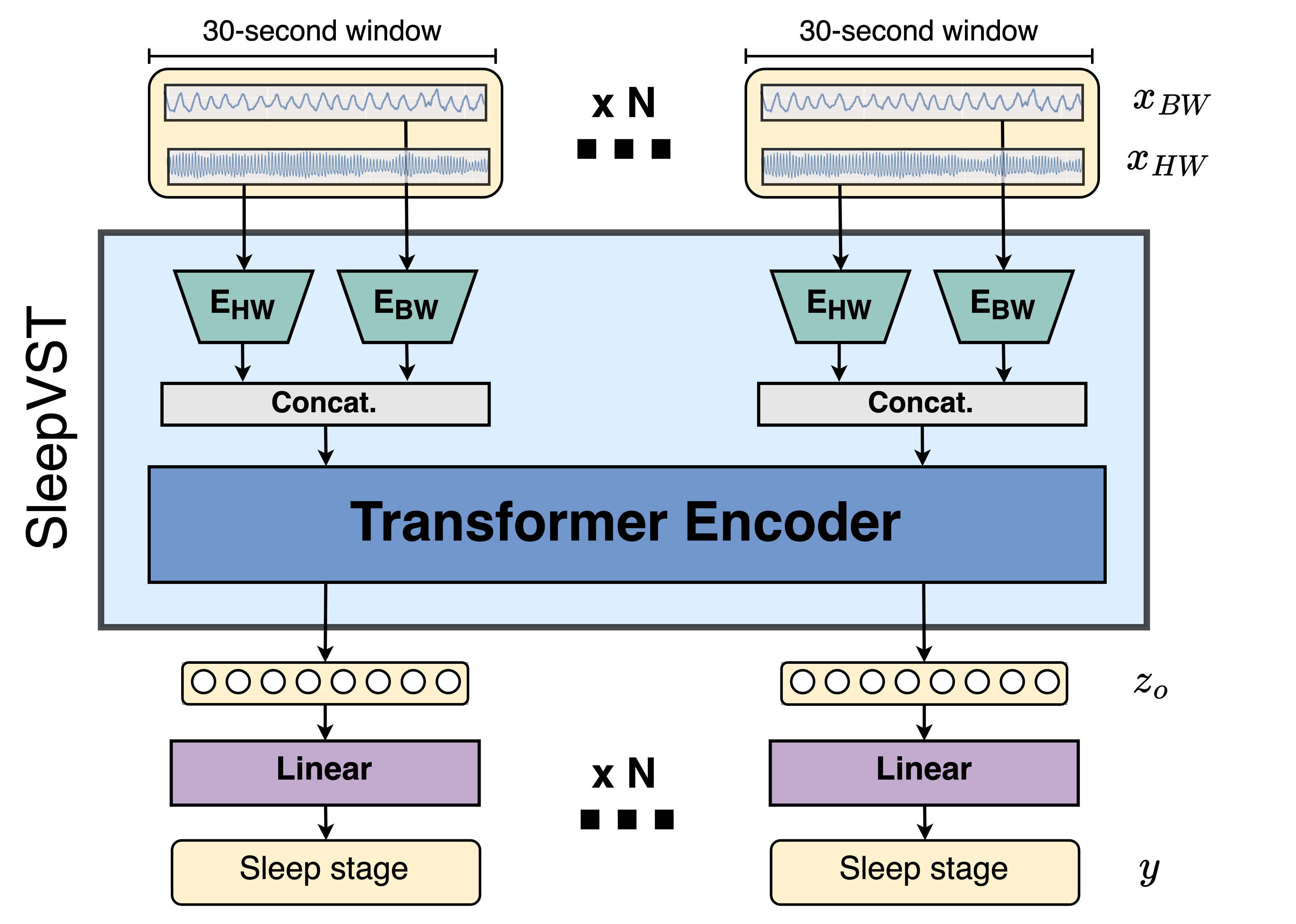}
   \caption{\textbf{SleepVST architecture.} Each 30-second window of heart ($x_{HW}$) and breathing waveforms ($x_{BW}$) is passed to a \textbf{\textcolor[HTML]{5fa27a}{patch encoder}}, which turns them into patch-level features. These features are concatenated and passed to a \textbf{\textcolor[HTML]{004C99}{transformer encoder}}. During pre-training, a \textbf{\textcolor[HTML]{A680B8}{linear layer}} turns the output feature sequence $z_o$ of length N from SleepVST into sleep stage classifications.}\label{fig:sleepvst}
\end{figure}

\Cref{fig:sleepvst} illustrates the SleepVST model architecture, which turns cardiac and respiratory waveforms into sequences of feature vectors using a transformer encoder. These feature vectors are then used to classify each 30-second sleep epoch. During pre-training, we used a linear layer for classification. When transferring to video data, we trained a new classifier that additionally used motion features derived from video as inputs, discussed further in \Cref{section:motion}.

\textbf{Patchification.}
The continuous heart and breathing waveforms (HW and BW) are divided into N windows i.e. patches, drawing on ideas from PatchTST~\citeMain{nie_time_2022}. We used non-overlapping 30-second windows since this is the interval at which sleep stages are classified according to AASM guidelines. We denote the patchified inputs as $\bm{x}_{HW} \in \mathbb{R}^{300\times N}$ and $\bm{x}_{BW} \in \mathbb{R}^{150\times N}$ respectively, where N is the sequence length.

\textbf{Encodings.}
Each waveform patch is normalised to zero mean and unit variance before being passed to encoding layers $E_{HW}$ and $E_{BW}$, which transform them into feature vectors $\bm{z}_{HW} \in \mathbb{R}^{D_{HW}\times N}$ and $\bm{z}_{BW} \in \mathbb{R}^{D_{BW}\times N}$. The feature vectors from each patch are then concatenated, producing a sequence of feature vectors $\bm{z}_i \in \mathbb{R}^{(D_{HW} + D_{BW})\times N}$ which are passed to a transformer encoder.

\begin{figure}[tb]
  \centering
   \includegraphics[width=0.9\linewidth]{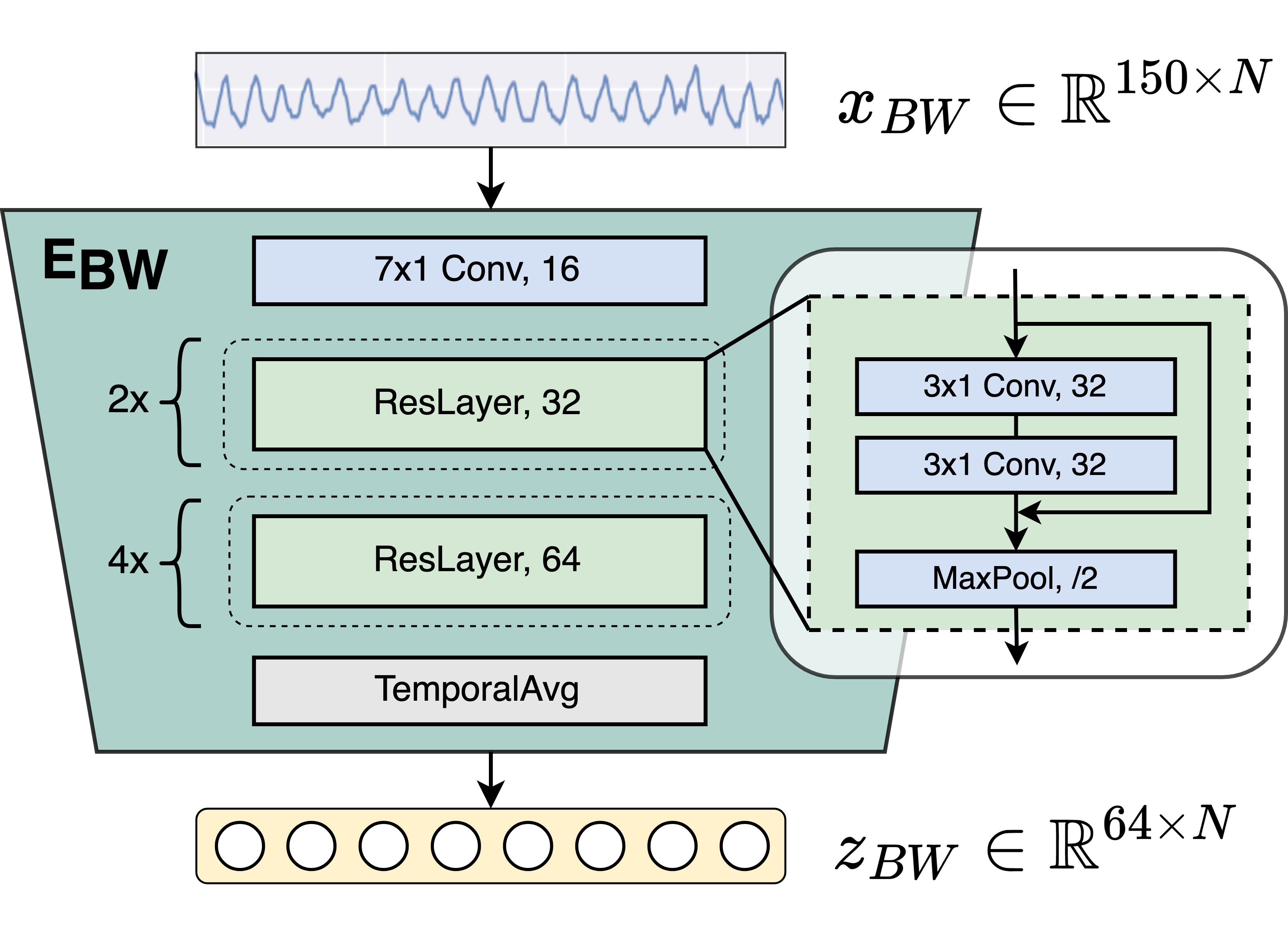}
   \caption{\textbf{Waveform encoder design.} Using a series of convolutional layers, the encoder turns sequences of signal patches into sequences of lower-dimensional feature vectors.}\label{fig:encoder}
\end{figure}

We used identical, shallow 1D ResNet-style~\citeMain{he_deep_2016} models for $E_{HW}$ and $E_{BW}$, illustrated in \Cref{fig:encoder}. This convolutional encoder design~\citeMain{baevski_wav2vec_2020} compresses the waveform patches into lower-dimensional feature vectors. This allowed us to keep $D_{HW}$ and $D_{BW}$ small, reducing the computational and memory complexity of the downstream transformer and the likelihood of over-fitting, particularly when transferring the model to the smaller video dataset.

This encoder design additionally provides a high degree of flexibility at the input level, such as the ability to use input signals of differing frequencies. In later ablation studies, this enables us to compare the performance of the SleepVST model with simpler variants which use 30-second patches of 1 Hz heart rate $\bm{x}_{HR}\in\mathbb{R}^{30\times N}$ and/or breathing rate $\bm{x}_{BR}\in\mathbb{R}^{30\times N}$ time-series as inputs, rather than underlying waveforms. When using either of these as inputs, we used linear projections~\citeMain{dosovitskiy_image_2021, nie_time_2022} for the input encoders, because of the lower dimensionality of $\bm{x}_{HR}$ and $\bm{x}_{BR}$. Throughout our experiments we use the same feature dimension i.e. $D_{HW}=D_{BW}=D_{HR}=D_{BR}=64$.

\textbf{Transformer layer.}
We used a Transformer encoder~\citeMain{vaswani_attention_2017} with sinusoidal position encodings to turn the sequence of input features $\bm{z}_i$ into an output sequence of features $\bm{z}_o$ with an identical feature dimension. Normal sleep exhibits long-range structures, known as sleep cycles~\citeMain{patel_physiology_2022}, which typically last around 90--120 minutes and typically occur four to five times per night. From its superior performance in many sequence modelling tasks, including EEG-based sleep staging~\citeMain{phan_sleeptransformer_2022}, we expect that the inductive biases of the transformer model (i.e. the attention mechanism) can enable it to learn to use relative position within a sleep cycle to classify sleep stages. 

\textbf{Classification.}
When training and testing on contact sensor data, we used a linear classification layer $\bm{W}_C \in \mathbb{R}^{C \times (D_{HW} + D_{BW})}$ to transform the sequence of output feature vectors $\bm{z}_o$ into a sequence of sleep stage probabilities $\bm{y} \in \mathbb{R}^{C\times N}$, where the number of classes $C$ depends on the sleep classification strategy used. In line with prior work, we classified sleep stages into four classes during pre-training: Wake--N1/N2--N3--REM, i.e. $C$~=~4.

\textbf{Model training and inference.}
In all experiments, we used a sequence length of N = 240 windows i.e. two hours. During training, we created batches by sampling sequences of length N with a step of 5 minutes from within each recording. The model was trained to minimise the multi-class cross-entropy loss using the AdamW~\citeMain{loshchilov_decoupled_2019} optimiser with a learning rate of 0.0005 and weight decay of 0.001. Other hyper-parameters, such as the no. encoder layers, are detailed in \S\ref{section:implementation}. To apply the model to arbitrary length input sequences during evaluation, we re-applied it at regular intervals. More details on this process are provided in \S\ref{section:tiling}.

\subsection{Motion features from Video}\label{section:motion}
Following prior work~\citeMain{nochino_sleep_2019}, we seek to produce motion features from video which encode for information which is predictive of the sleep stage, such as the rate and distribution of body movements~\citeMain{wilde-frenz_rate_1983}. To do so, we define a set of parameterised time-series features that collectively quantify motion over a range of spatial and temporal scales.

\begin{figure}[htb]
  \centering
   \includegraphics[width=0.95\linewidth]{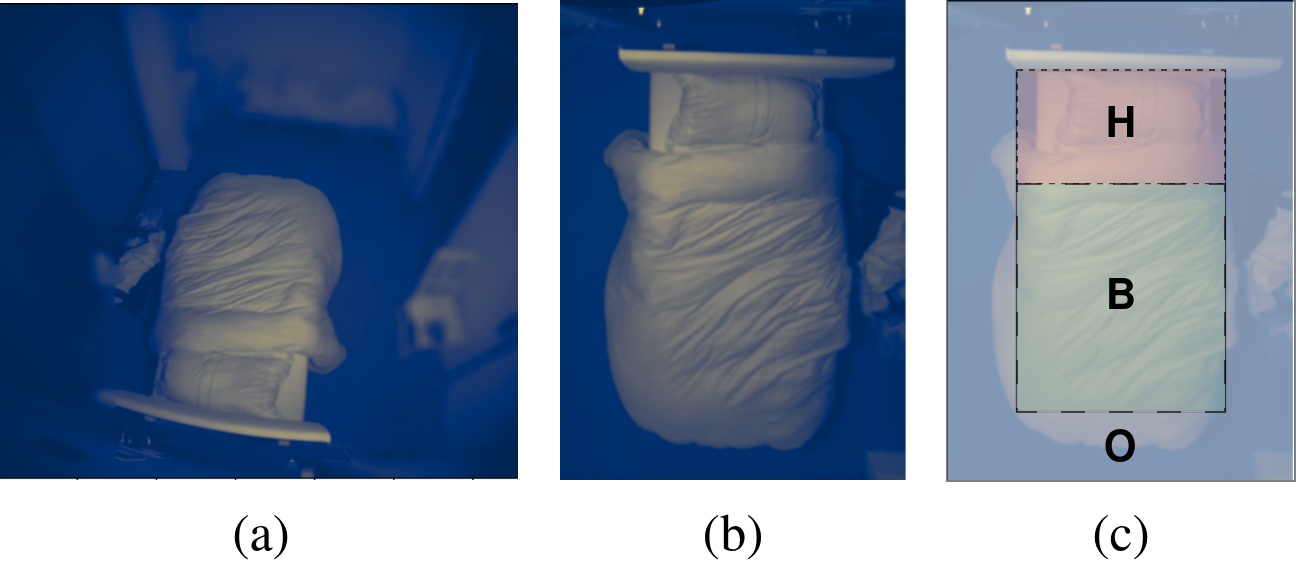}
   \caption{\textbf{Example processing of an NIR video frame from the OSV dataset.} (a) Real viewpoint. (b) Virtual viewpoint. (c) Head (H), body (B), and outer (O) bed regions. The distance from the camera to the head region is around 1.5 m.}\label{fig:flowregions}
\end{figure}

\textbf{Optical flow estimation.} We first estimated a 2D optical flow field $\bm{u}(t, x, y)\in \mathbb{R}^2$ over the bed region. Inspired by sleep pose detection work~\citeMain{mohammadi_transfer_2021}, we applied a homography transformation~\citeMain{hartley_multiple_2003} to the individual video frames before estimating the flow field, to achieve camera viewpoint invariance (shown in \Cref{fig:flowregions}).

\textbf{Feature definitions.} Given an arbitrary region R of the optical flow field, we define the following parametrised signals:
\begin{align} 
v(t;R)&=\max_{(x,y)\in R} |\bm{u}(t, x, y)| \label{equation:maxsignal}\\ 
s(t;R)&=\frac{1}{|R|}\sum_{(x,y)\in R} |\bm{u}(t, x, y)|\label{equation:sumsignal}
\end{align}
i.e. the maximum \eqref{equation:maxsignal} and the average \eqref{equation:sumsignal} of the optical flow magnitude in the region R at time $t$. Using these signals, we define two time-series features that quantify motion within a time window $\Delta$ up to time $t$:
\begin{align} 
f_1(t; R, \Delta) &= \quad\sum_{t-\Delta}^{t} v(t; R) \\ 
f_2(t; R, \Delta) &= \quad\sum_{t-\Delta}^{t} s(t; R)
\end{align}
Plus two additional features which measure the time elapsed since either signal went above a threshold $\delta$:
\begin{equation}
\begin{aligned}
f_3(t; R, \delta) & = \sum_{t-\tau}^{t} 1\\
\textrm{where} \quad \tau & = \max_{t' \leq t} t' \\
\textrm{s.t.} \quad & v(t'; R) > \delta\\
\end{aligned}
\end{equation}
\begin{equation}
\begin{aligned}
f_4(t; R, \delta) & = \sum_{t-\tau}^{t} 1\\
\textrm{where} \quad \tau & = \max_{t' \leq t} t' \\
\textrm{s.t.} \quad & s(t'; R) > \delta\\
\end{aligned}
\end{equation}

We calculated each feature for all possible combinations of parameters $\Delta \in \{30, 300\}$ seconds, $\delta \in \{0.01, 0.1, 1\}$ and $R \in \{H, B, O\}$, where $\{H, B, O\}$ are approximate head, body and outer bed regions as illustrated in \Cref{fig:flowregions}. This resulted in ($2$+$2$+$3$+$3$) time-series features per region, i.e. 30 unique time-series features. Additionally, we provided the video-based classifier (see \Cref{section:videotransfer}) with time-shifted features $f_i(t+T)$ for $T \in \{\text{-}90, 0, \text{+}90\}$ seconds. This gave a total of 90 motion features for each sleep epoch.

\subsection{Transferring SleepVST to Video}\label{section:videotransfer}
To transfer the SleepVST model to video using the OSV dataset, we removed the linear classification layer learnt during pre-training and used the model as a frozen feature extractor, applying it to sequences of cardiac and respiratory waveforms measured from video to produce sequences of output features $\bm{z}_o$. We then trained a new classifier that used these features, plus the motion features derived in the previous section, to classify individual sleep epochs. This allowed us to leverage the representational capacity of the transformer model, by pre-training on the much larger contact-sensor datasets, whilst also enabling us to incorporate motion information from the video dataset. We followed the design choice of~\citeMain{carter_deep_2023} and used a Random Forest~\citeMain{breiman_random_2001} for our classifier due to its low computational cost and robustness to overfitting.
\section{Experimental Results}
Sleep staging performance has commonly been reported in terms of classification accuracy ($\text{Acc}_T$) and total Cohen's kappa ($\kappa_T$) over all sleep epochs in the dataset, and/or using per-night mean statistics, $\text{Acc}_\mu$ and $\kappa_\mu$, i.e.:
\begin{align}
    \kappa_T &= K(\sum_{i=1}^{N}  c(\bm{y}_i, \hat{\bm{y}}_i)) \label{equation:kappa_total}\\
    \kappa_\mu &= \frac{1}{N} \sum_{i=1}^{N} K ( c(\bm{y}_i, \hat{\bm{y}}_i))\label{equation:kappa_mu}
\end{align}
\begin{figure*}[!tb]
  \centering
   \includegraphics[width=0.99\linewidth]{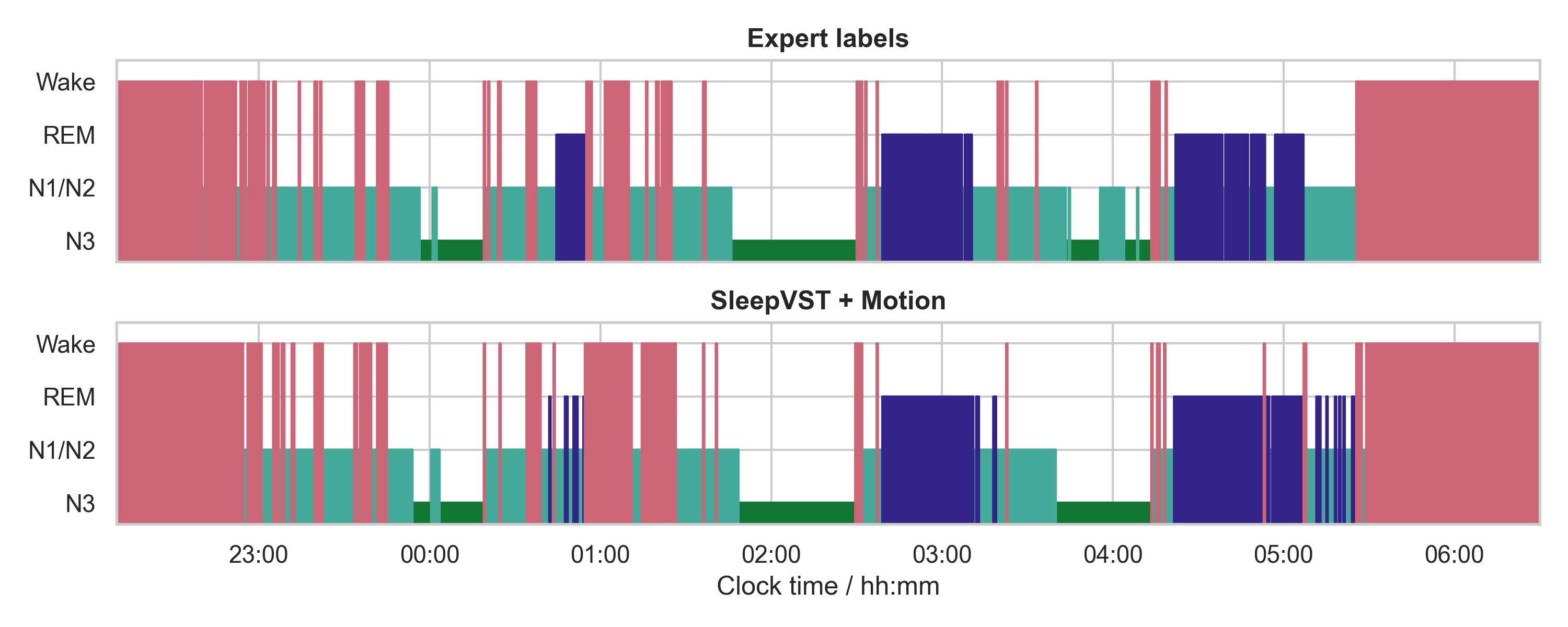}
   \caption{\textbf{Example four-class sleep hypnograms from the OSV dataset.} (top) Scored by an expert using signals from the vPSG recording. (bottom) Automatically generated from near-infrared video using SleepVST. (Cohen's $\kappa=0.71$ between model and expert sleep stages.)}\label{fig:middlehyp}
\end{figure*}
where $\bm{y}_i, \hat{\bm{y}}_i \in \mathbb{R}^{C\times L_i}$ are the one-hot encoded model and scorer labels for recording $i$ of length $L_i$ and $c(\bm{y}_i, \hat{\bm{y}}_i) \in \mathbb{R}^{C\times C}$ is the confusion matrix between them. $K(\cdot)$ is the function which calculates the Cohen's kappa statistic~\citeMain{cohen_coefficient_1960} from a confusion matrix and is defined in \S\ref{section:kappa_definition}. We report all of these statistics, to make fairer comparisons with prior work, and to aid future comparisons.

\subsection{Pre-training on Contact Sensors}
From the SHHS dataset, we randomly selected 500 participants who participated in both visits, to form a test set of 1000 nights of data. This ensured that no participant appeared in both the training/validation and test partitions. Our test set is listed in \S\ref{section:shhs_ids}. From the MESA dataset, we used the same test set of 204 nights as Kotzen \etal~\citeMain{kotzen_sleepppg-net_2023}. The remaining nights from both the SHHS and MESA datasets were pooled together and randomly split into training and validation sets using a 75:25 split.
\begin{table*}[htb]
    \begin{center}
    \caption{Comparison of four-class cardio-respiratory sleep staging methods using contact sensors.}
    \small
    \begin{tabular}{@{}llllllll@{}}
        \toprule
             &      &         &   & \multicolumn{2}{c}{Cohen's $\kappa$} & \multicolumn{2}{c}{Accuracy / \%}  \\ \cmidrule(l){5-6}\cmidrule(l){7-8}
        Dataset & N & Method                                             & Modalities         & $\kappa_\mu$  & $\kappa_T$ & $\text{Acc}_\mu$ & $\text{Acc}_T$\\ \midrule
        SHHS~\citeMain{quan_sleep_1997}    & 296  & Bakker et al.~\citeMain{bakker_estimating_2021}   & HR+Thor$^*$+Air$^\dag$ & -  & 0.64 & - & 76.7 \\
           &  800   &Sridhar et al.~\citeMain{sridhar_deep_2020}    & HR (ECG) & 0.65 & 0.66 & 77.3 & 77.0\\
          & 1000 & SleepVST     & ECG+Thor$^*$ & \textbf{0.73}  & \textbf{0.75} & \textbf{82.8} & \textbf{83.0}   \\ \midrule[0.5pt]
        MESA~\citeMain{chen_racialethnic_2015}   &  296 & Bakker et al.~\citeMain{bakker_estimating_2021}   & HR+Thor$^*$+Air$^\dag$     & - & 0.68 & - & 79.8 \\
          & 194   & Sridhar et al.~\citeMain{sridhar_deep_2020}       & HR (ECG)  & -  & 0.69 & - & 80.0\\
          & 204     & Kotzen et al.~\citeMain{kotzen_sleepppg-net_2023} & PPG  & -  & 0.73$^\ddag$ & - & 82.6$^\ddag$ \\
           & 204   & SleepVST                                           & ECG+Thor$^*$              & \textbf{0.76}  & \textbf{0.77}    & \textbf{85.1} & \textbf{85.2} \\ \bottomrule
           \multicolumn{6}{l}{\scriptsize{$^*$Thoracic respiratory effort.\, $^\dag$Nasal airflow.\, $^\ddag$From reported confusion matrices.}}   
    \end{tabular}\label{table:contactcomparison}
    \end{center}
\end{table*}

\Cref{table:contactcomparison} compares the overall performance of SleepVST with prior methods for cardio-respiratory sleep staging from contact sensors. These have used input modalities such as the photoplethysmogram (PPG) waveform and the ECG-derived heart rate. Using the ECG and thoracic respiratory signals, SleepVST outperforms prior methods for sleep staging from cardio-respiratory signals on both datasets.

\subsection{Video-based Sleep Staging}
In this section, we report the performance of the SleepVST model after transfer to the OSV video dataset. We follow the evaluation procedure of~\citeMain{carter_deep_2023}, training and evaluating on the 50 nights of data using 10-fold cross-validation with non-overlapping folds each containing five nights of data.

In \Cref{table:videocomparison}, we see that our method achieves significantly better performance in terms of both accuracy and Cohen's $\kappa$ statistic when compared with prior video-based methods evaluated on similar study populations.
\begin{table*}[htb]
    \begin{center}
    \caption{Comparison of four-class sleep staging methods using video cameras.}
    \small
    \begin{tabular}{@{}llllllll@{}}
        \toprule
          &    &        &  & \multicolumn{2}{c}{Cohen's $\kappa$} & \multicolumn{2}{c}{Accuracy / \%}  \\ \cmidrule(l){5-6}\cmidrule(l){7-8}
        Dataset & N & Method & Modalities$^{**}$  & $\kappa_\mu$ & $\kappa_T$ & $\text{Acc}_\mu$ & $\text{Acc}_T$  \\ \midrule
        Healthy infants~\citeMain{kamon_development_2022} & 8 & Kamon et al.~\citeMain{kamon_development_2022}   & Motion & 0.26 & - & 48.0 & -\\
        Healthy adults~\citeMain{nochino_sleep_2019} & 6 & Nochino et al.~\citeMain{nochino_sleep_2019}   & Motion & 0.19 & - & 40.5 & -\\
        HealthBed Study~\citeMain{van_meulen_contactless_2023} & 46 & van Meulen et al.~\citeMain{van_meulen_contactless_2023} & HW$^{*}$ & 0.49 & 0.49$^\ddag$ & 67.9 & 67.9$^\ddag$ \\
        Oxford Sleep Volunteers & 50 & Carter et al.~\citeMain{carter_deep_2023}  & HR+BR+Motion & 0.61 & 0.64$^\ddag$ & 73.4 & 74.3$^\ddag$\\ 
         & 50 & SleepVST  & HW$^{*}$+BW$^\dag$+Motion &  \textbf{0.68}   & \textbf{0.71} & \textbf{77.7} & \textbf{78.8}\\ \bottomrule
          \multicolumn{8}{l}{\scriptsize{$^{**}$Camera-derived. $^{*}$Cardiac pulse (heart) waveforms. $^\dag$Respiration (breathing) waveforms. $^\ddag$From reported confusion matrices.}}
    \end{tabular}\label{table:videocomparison}
    \end{center}
\end{table*}

\begin{figure}[!htb]
  \centering
   \includegraphics[width=0.99\linewidth]{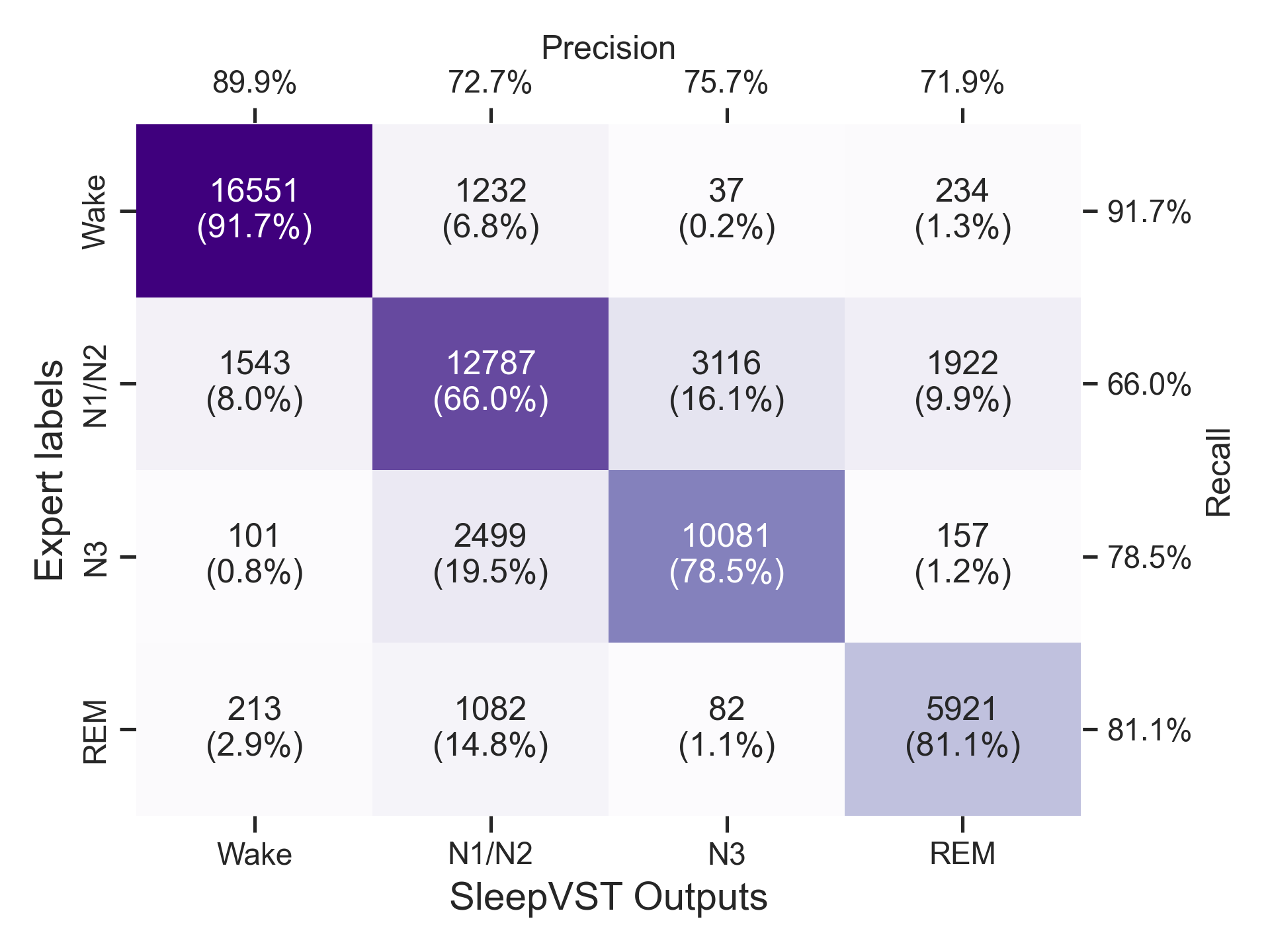}
   \caption{Confusion matrix between SleepVST and expert labels on four-class video based sleep staging.}\label{fig:cmat}
\end{figure}

\Cref{fig:cmat} shows the total test confusion matrix between model and human expert (polysomnographer) classifications on the OSV dataset. We observe particularly high accuracy in distinguishing Wake and REM sleep stages.

\begin{figure}[!htb]
  \centering
   \includegraphics[width=0.99\linewidth]{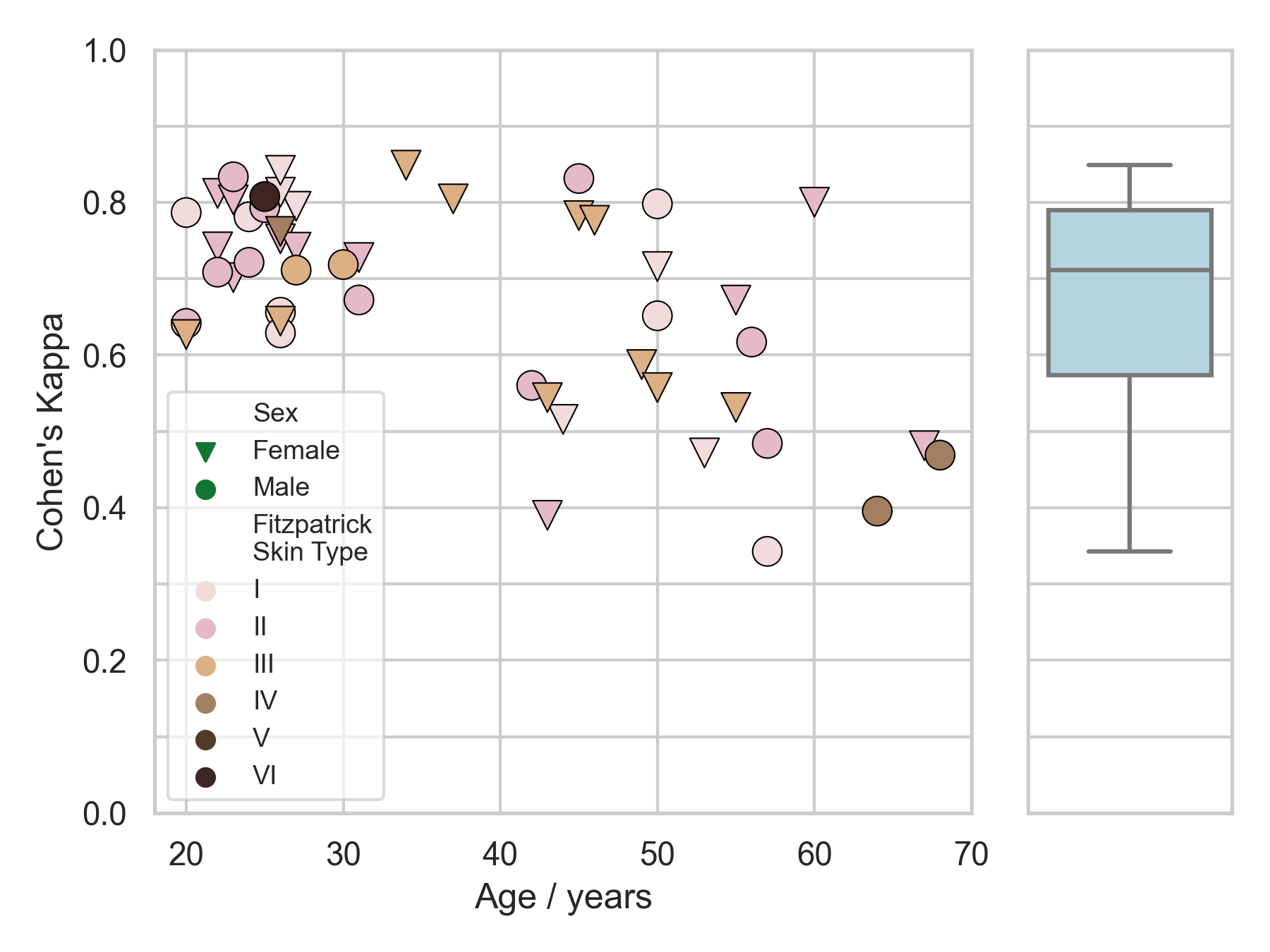}
   \caption{Scatter and box plot of Cohen’s $\kappa$ against age, sex and Fitzpatrick skin type$^\text{\ref{fitzfootnote}}$ for individuals from the OSV dataset.}\label{fig:agescatter}
\end{figure}

\textbf{Population performance.}
\Cref{fig:agescatter} shows the distribution of Cohen's $\kappa$ values across participants from the OSV dataset plotted against age, sex and Fitzpatrick skin type\footnote{Plotted using representative skin tones from \citeMain{ward_clinical_2017}.\label{fitzfootnote}}~\citeMain{fitzpatrick_validity_1988}. We observe a decrease in performance with age, as observed in prior sleep staging work~\citeMain{sun_sleep_2020, carter_deep_2023}. In older adults, measures of cardio-respiratory activity (e.g. heart rate variability) are known to decline with age~\citeMain{ziegler_assessment_1992}. This likely reduces the distance between sleep stages in cardio-respiratory input space and thus makes them harder to distinguish. Additional video data from older subjects would likely help to improve this, which is discussed further in \S\ref{section:additionalresults}.

\textbf{Example SleepVST output.}
\Cref{fig:middlehyp} shows an example hypnogram generated entirely from near-infrared video using our method. This example corresponds to the median Cohen's $\kappa$ between model and expert from the OSV dataset. The model correctly identifies all three REM cycles, periods of deep sleep, and brief awakenings throughout the night. Additional hypnogram examples can be found in \S\ref{section:additionalresults}.

\subsection{Ablation Studies}
\textbf{Video transfer strategy.}
In \Cref{table:transfermethod}, we compare the performance of three methods for transferring the SleepVST model to video data:
\begin{enumerate}
    \item Directly applying the pre-trained SleepVST model to cardiac and respiratory waveforms from video data.
    \item Using SleepVST as a feature extractor and:
        \begin{enumerate}
            \item training a new classification layer without using motion features.
            \item training a new classification layer that also uses motion features i.e. our best-performing approach.
        \end{enumerate}
\end{enumerate}
Here we see that training a new classifier for video data is essential to the effectiveness of our approach. This step likely allows the classifier to ignore parts of the learnt feature space that do not transfer between contact-sensor and video-derived waveforms i.e. sensor-specific features.
\begin{table}[htb]
\centering
\caption{SleepVST video-based sleep staging performance for different model transfer strategies.}
\small
\begin{tabular}{@{}lllll@{}}
\toprule
  & $\kappa_\mu$ & $\kappa_T$ & $\text{Acc}_\mu$ / \% & $\text{Acc}_T$ / \% \\ \midrule
Direct apply   & 0.510 & 0.536 & 66.7 & 67.3\\
Transfer w/o motion & 0.660  & 0.689 & 76.4 & 77.4\\
Transfer w/ motion & \textbf{0.677} &  \textbf{0.708}  & \textbf{77.7} & \textbf{78.8}\\\bottomrule
\end{tabular}
\label{table:transfermethod}
\end{table}
\begin{figure}[htb]
  \centering
   \includegraphics[width=1\linewidth]{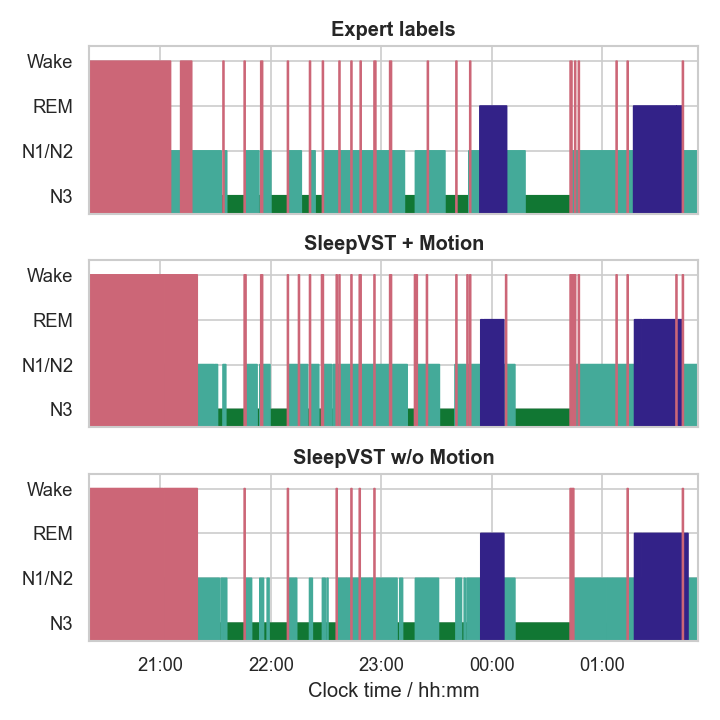}
   \caption{\textbf{Example four-class hypnograms during fragmented sleep from the OSV dataset.} The use of motion features improves accuracy during short awakenings between 22.00 and 00.00.}\label{fig:motionhyp}
\end{figure}

\textbf{Motion features.}
\Cref{table:motionablation} reports the effectiveness of the two main components of our motion feature set: the set of feature definitions $f_i$ and the set of regions used $R$. Additional ablations can be found in \S\ref{section:additionalresults}. Here we define and report an additional metric $\kappa_F$ which is the Cohen's $\kappa$ statistic computed over epochs near transitions to/from Wake ($W$) i.e. where $\hat{y}_i^{t}\neq\hat{y}_i^{t+\Delta}=W$ or $\hat{y}_i^{t}=W\neq\hat{y}_i^{t+\Delta}$ for any $\Delta\in[-2,2]$ epochs. This metric better quantifies accuracy during fragmented sleep, such as in \Cref{fig:motionhyp}.

As measured by $\kappa_F$, we see that motion features aid classification during fragmented sleep. The ability to accurately measure sleep fragmentation is particularly important since it is linked with daytime sleepiness and decreased cognitive performance~\citeMain{martin_effect_1996}.
\begin{table}[htb]
\centering
\caption{Ablation study of motion feature parameters.}
\small
\begin{tabular}{@{}lllll@{}}
\toprule[1pt]
 & & & \multicolumn{2}{c}{Cohen's $\kappa$}\\\cmidrule(l){4-5}
Ablation    & Parameter Set & $n_f^*$ & $\kappa_T$ & $\kappa_F$ \\ \midrule
Feature set & $f_i\in\{\}$ & 0 & 0.689 & 0.397                         \\
    & $f_i \in \{f_1, f_2\}$ & 36  & 0.697 & 0.467                  \\ 
    & $f_i \in \{f_3, f_4\}$ & 54  & 0.704 & 0.465                  \\
    & $f_i \in \{f_1, f_2, f_3, f_4\}$ & 90 & \textbf{0.708} &  \textbf{0.491}      \\ \midrule[0.25pt]
Region(s) & $R \in \{H \cup B\}$ & 30 & 0.702 & 0.461                         \\
          & $R \in \{H \cup B, O\}$ & 60 & 0.704 & 0.468              \\
          & $R \in \{H, B, O\}$ & 90 & \textbf{0.708}  &  \textbf{0.491} \\ \bottomrule
{\scriptsize $^*$No. features.}
\end{tabular}
\label{table:motionablation}
\end{table}

\textbf{Cardio-respiratory inputs.}
In \Cref{table:vstinputs}, we compare the performance of the SleepVST model with variants which use heart rate $\bm{x}_{HR}$ or breathing rate $\bm{x}_{BR}$ as inputs instead of the underlying waveforms, as described in \Cref{section:vst}. We see that the use of waveforms, rather than derived heart and breathing rates, leads to improvements in sleep staging accuracy across all datasets.
\begin{table}[htb]
\centering
\caption{Comparison of four-class sleep staging performance for different cardio-respiratory input combinations to SleepVST.}
\small
\begin{tabular}{@{}llll@{}}
\toprule
 & \multicolumn{3}{c}{Cohen's $\kappa_T$}\\\cmidrule(l){2-4}
Input signals & SHHS & MESA & OSV\\ \midrule
$\bm{x}_{HR}\:,\,\bm{x}_{BR}$ & 0.683 & 0.722 & 0.681\\
$\bm{x}_{HW},\,\bm{x}_{BR}$ & 0.718 & 0.736 & 0.684\\
$\bm{x}_{HR}\:,\,\bm{x}_{BW}$ & 0.723 & 0.758 & 0.702\\
$\bm{x}_{HW},\,\bm{x}_{BW}$ & \textbf{0.749} & \textbf{0.765} & \textbf{0.708}\\\bottomrule
\end{tabular}
\label{table:vstinputs}
\end{table}
\section{Conclusions}
We have introduced SleepVST, a transformer model for sleep stage classification from cardio-respiratory waveforms. Using a transfer learning approach, we have shown how SleepVST can be used to perform sleep stage classification entirely from near-infrared video.

Our results advance the state-of-the-art in video-based sleep staging, narrowing the gap to expert-level performance and taking a significant step towards important clinical applications. More broadly, our work shows that a model pre-trained on a task using cardio-respiratory waveforms from large contact-sensor datasets can be effectively transferred to waveforms measured from video. This approach can benefit wider applications such as the detection of arrhythmias and sleep apnoeas, from cardiac and respiratory waveforms respectively.

Finally, we believe there is important future work in learning more expressive representations of motion during sleep. In addition to improving the performance of video-based sleep staging, this could also enable the detection of sleep movement disorders, such as REM behaviour disorder, removing the existing need for time-consuming, manual video review~\citeMain{stefani_sleep_2020}.

{
\small 

\section*{Acknowledgements}
This work was supported by the EPSRC Centre for Doctoral Training in Autonomous Intelligent Machines and Systems [EP/S024050/1] and funded by Oxehealth Ltd. Figure 1 was created with BioRender.com. We kindly thank the National Sleep Research Resource for granting access to SHHS and MESA.

}
{
    \small
    \bibliographystyle{ieeenat_fullname}
    \bibliography{main}
}
\appendix
\clearpage
\maketitlesupplementary

\section{Broader Impact Statement}
\label{section:impact}
Video data is a sensitive modality and its use in healthcare must come in conjunction with patient consent and clinical assessment. In sleep medicine, video polysomnography is considered the gold-standard monitoring technique, and the use of video enables accurate diagnoses of specific conditions such as REM behaviour disorder~\citeSupp{stefani_sleep_2020}. For wider applications, the algorithms introduced in this work have been designed such that the video data can be processed into intermediate motion and cardio-respiratory signals for further processing offline. This means it is possible for camera data to be processed in real-time without being stored.

Machine learning methods have the potential to make a transformative impact in healthcare. However, they should be appropriately evaluated across diverse populations and in real-world scenarios, to understand their limitations, mitigate potential biases, and reduce clinical risk. In future work, we plan to evaluate the performance of our method across a broader population, including a greater number of older individuals and individuals with darker skin tones, and individuals with diagnosed sleep disorders.

\section{Oxford Sleep Volunteers Dataset}
\label{section:appendix_data}
\begin{figure}[htb]
  \centering
   \includegraphics[width=0.95\linewidth]{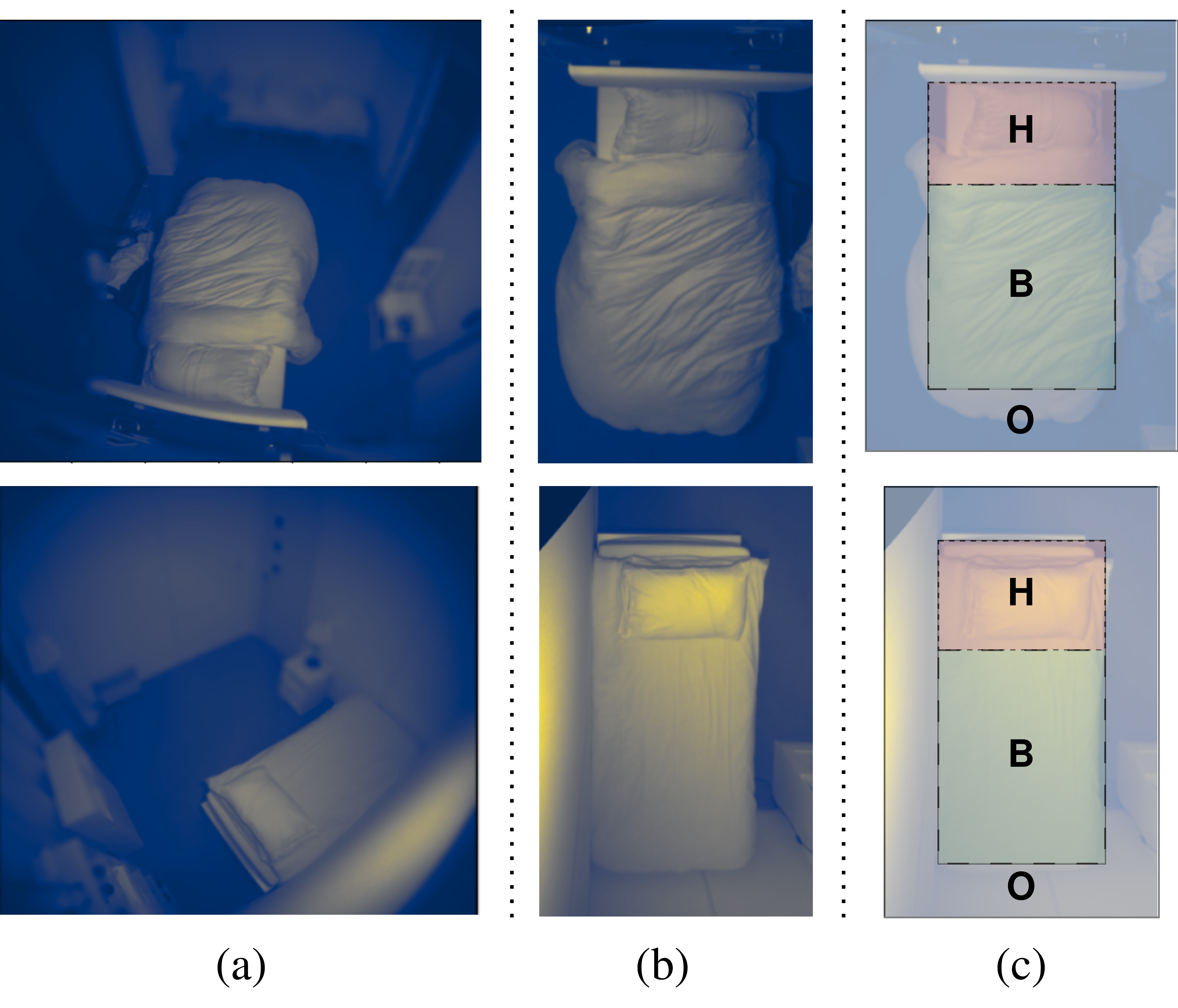}
   \caption{\textbf{Example processing of near-infrared video frames from the OSV dataset.} (a) Real viewpoints and bed positions for each room. (b) Cropped, virtual viewpoints obtained using homography transformations. (c) Head (H), body (B), and outer (O) bed regions.}\label{fig:flowregions_demo}
\end{figure}

Throughout this work, we used the video polysomnography dataset introduced by Carter \etal~\citeSupp{carter_deep_2023} for our video-based sleep staging experiments, which we refer to as the Oxford Sleep Volunteers (OSV) dataset. Figure \ref{fig:flowregions_demo} shows the two camera viewpoints and room layouts present in the dataset, along with transformed frames, and approximate head, body and outer bed regions. The homography transformations and regions were manually determined from the known camera parameters and room geometry. For more information on the dataset, including population demographics and room geometries, we refer to the original work~\citeSupp{carter_deep_2023}.

\section{Additional SleepVST Results}
\label{section:additionalresults}
\textbf{Additional model hypnograms.}
Additional examples of sleep hypnograms generated using SleepVST from video data are shown in Figure \ref{fig:extrahyps}.

\begin{figure}[htb]
  \centering
   \includegraphics[width=1\linewidth]{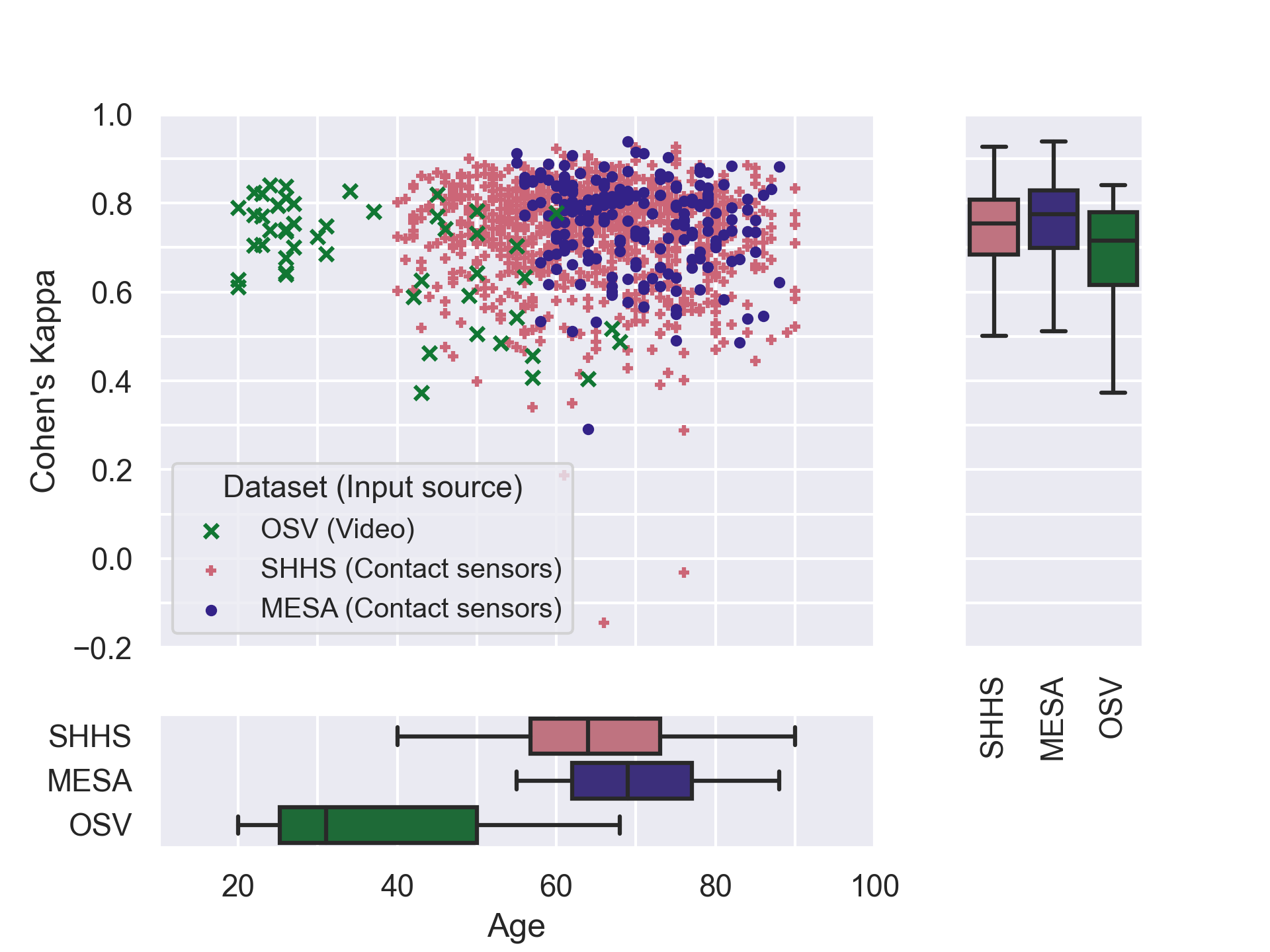}
   \caption{Scatter and box plots of Cohen’s $\kappa$ and age distributions across the SHHS, MESA and OSV datasets.}\label{fig:agescatter_all}
\end{figure}

\textbf{Cohen's $\kappa$ distribution across datasets.}
\Cref{fig:agescatter_all} shows the distribution of Cohen's $\kappa$ values with age across all three datasets. Despite being trained on contact sensor waveforms from much older subjects in the SHHS and MESA datasets, SleepVST successfully transfers to video-derived waveforms, nearly reaching parity with performance using contact sensors. This highlights the effectiveness and generality of the learnt feature space. Given the well-known changes in autonomic activity with age~\citeSupp{ziegler_assessment_1992}, pre-training with additional data from younger subjects may further improve transfer learning performance on the existing dataset. As may fine-tuning the entire SleepVST network on video data, rather than freezing the weights and using it as a fixed feature extractor. Conversely, more video data from older participants during the transfer learning phase would likely help to improve the performance for the existing older participants.

\begin{figure}[htb]
  \centering
   \includegraphics[width=0.95\linewidth]{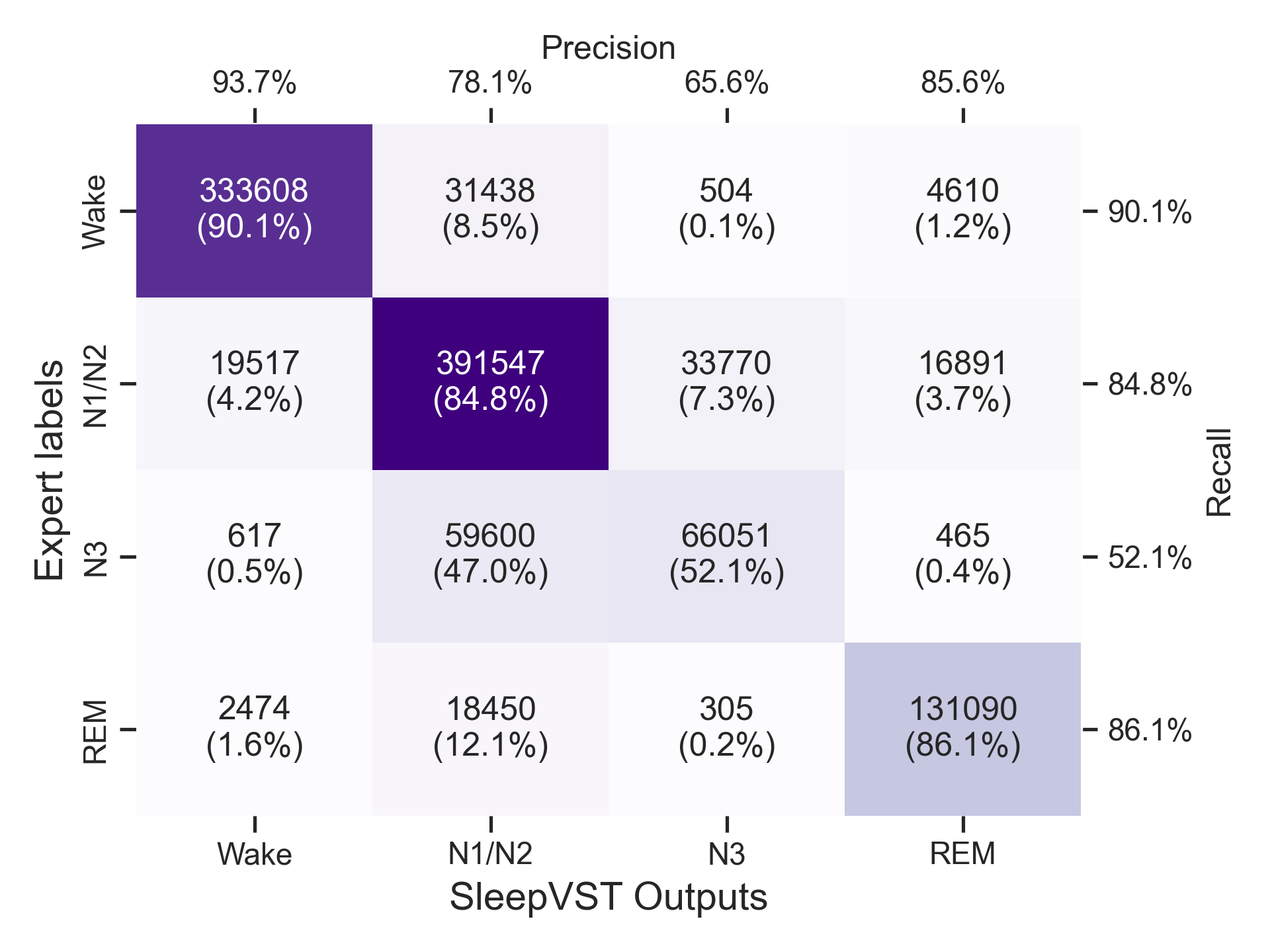}
   \caption{SleepVST confusion matrix with expert labels, evaluated on the SHHS test set.}\label{fig:cmat_shhs}
\end{figure}

\begin{figure}[htb]
  \centering
   \includegraphics[width=0.95\linewidth]{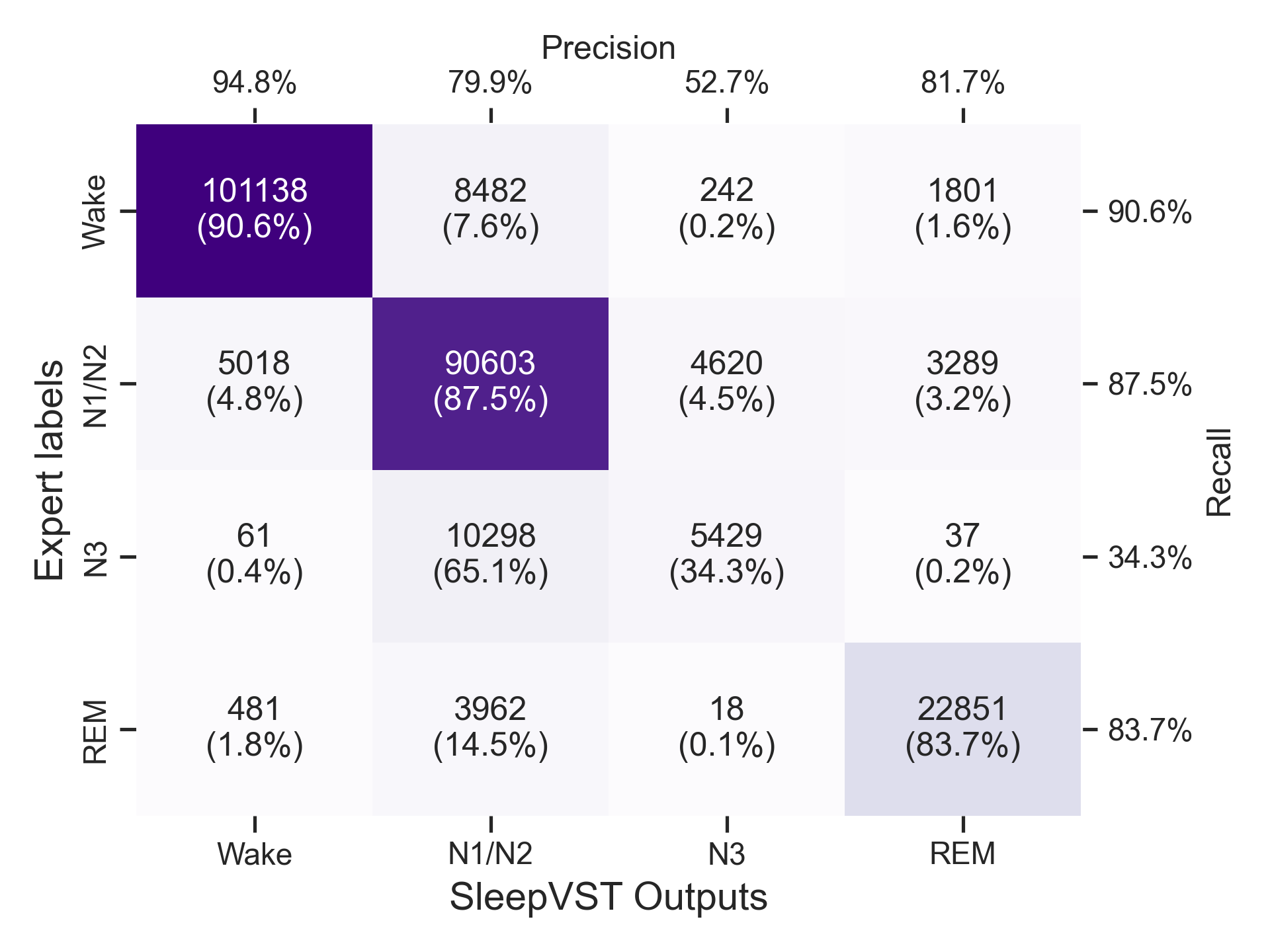}
   \caption{SleepVST confusion matrix against expert labels, evaluated on the MESA test set.}\label{fig:cmat_mesa}
\end{figure}

\textbf{Pre-training confusion matrices.} Figures \ref{fig:cmat_shhs} and \ref{fig:cmat_mesa} show the four-class sleep staging confusion matrices obtained after pre-training, for the SHHS and MESA test sets respectively. During pre-training, we chose to use an unbalanced cross-entropy loss. Using a class-weighted loss instead would likely help to reduce the observed rate of misclassification of (less frequent) N3 sleep.

\textbf{Varying classification strategy.}
In \Cref{table:classificationstrategy}, we report the performance of our method for different sleep stage classification strategies, to aid future comparisons. In each case, we use the same SleepVST model (pre-trained on four-class sleep staging) as a feature extractor, and train a new classifier using video data.

\begin{table}[htb]
\centering
\caption{Video-based sleep staging performance using SleepVST with different classification strategies.}
\small
\begin{tabular}{@{}lllll@{}}
\toprule
              & $\kappa_{\mu}$ & $\kappa_T$  & $\text{Acc}_\mu$ / \% & $\text{Acc}_T$ / \% \\ \midrule
\scriptsize{Sleep--Wake}   & 0.782{\footnotesize$\pm$0.197}& 0.857 & 93.3{\footnotesize$\pm$5.4}& 93.7  \\
\scriptsize{W--NREM--REM}   & 0.760{\footnotesize$\pm$0.139}& 0.795 & 87.4{\footnotesize$\pm$6.6}& 87.9  \\
\scriptsize{W--N1/N2--N3--REM} & 0.677{\footnotesize$\pm$0.133} & 0.708 & 77.7{\footnotesize$\pm$8.9} & 78.8\\
\scriptsize{W--N1--N2--N3--REM}  & 0.646{\footnotesize$\pm$0.139}& 0.674 & 74.1{\footnotesize$\pm$10.7} & 75.3 \\ \bottomrule
\end{tabular}
\label{table:classificationstrategy}
\end{table}

\textbf{Additional motion feature ablations.}
\Cref{table:motionablation_full} shows the effectiveness of the remaining components of our feature set. Higher thresholds ($\delta=1$) measure the elapsed time since large movements, which are more important for overall agreement $\kappa_T$. Lower thresholds ($\delta=0.01$) are more sensitive to smaller movements e.g. brief awakenings, leading to greater agreement around fragmented sleep $\kappa_F$.
\begin{table}[htb]
\centering
\caption{Ablation of motion feature parameters $T, \delta$ and $\Delta$.}
\small
\begin{tabular}{@{}lllll@{}}
\toprule[1pt]
 & & & \multicolumn{2}{c}{Cohen's $\kappa$}\\\cmidrule(l){4-5}
Ablation    & Parameter Set & $n_f^*$ & $\kappa_T$ & $\kappa_F$ \\ \midrule
Threshold(s) & $\delta \in \{0.01\}$ & 54 & 0.698 & 0.485 \\
             & $\delta \in \{0.1\}$ & 54 & 0.702 & 0.486 \\
             & $\delta \in \{1\}$& 54 & 0.705 & 0.471 \\
          & $\delta \in \{0.01, 0.1, 1\}$ & 90& \textbf{0.708}  &  \textbf{0.491} \\ \midrule[0.25pt]
Time shift(s) & $T \in \{0\}$ & 30 & 0.705 & 0.464 \\
          & $T \in \{\text{-}90, 0, 90\}$ & 90 & \textbf{0.708}  &  \textbf{0.491} \\ \midrule[0.25pt]
Window(s) & $\Delta \in \{30\}$ & 72 & 0.706 & 0.486\\
          & $\Delta \in \{300\}$ & 72 & 0.705 & 0.472\\
          & $\Delta \in \{30, 300\}$ & 90 & \textbf{0.708}  &  \textbf{0.491} \\ \bottomrule
          {\scriptsize $^*$No. features.}
\end{tabular}
\newline
\label{table:motionablation_full}
\end{table}

\section{SleepVST Architecture Details}
\label{section:implementation}
\textbf{Convolutional encoder.} Each convolutional layer, denoted `Kx1 Conv, M' in \Cref{fig:encoder}, consisted of 1D convolutions with kernel size K, stride 1, and M output channels, followed by batch normalisation~\citeSupp{ioffe_batch_2015}, and ReLU activation.

\textbf{Transformer encoder.}
The parameters of our transformer encoder are detailed in \Cref{table:transformer}, these values were informed by the original design of Vaswani \etal~\citeSupp{vaswani_attention_2017}. To improve training stability, we employed pre-layer normalisation~\citeSupp{xiong_layer_2020} within each transformer encoder layer.
\begin{table}[htb]
\centering
\caption{SleepVST transformer encoder parameters.}
\begin{tabular}{@{}ll@{}}
\toprule
Architecture Parameter             & Value\\ \midrule
Encoder layers & 6 \\
Self-attention heads & 8\\
Encoder dropout probability & 0.1\\
MLP size & 512\\\bottomrule
\end{tabular}
\label{table:transformer}
\end{table}

\textbf{Training.}
Each training run was performed using a single NVIDIA A10 GPU with 24 GB RAM. Using a sequence length of two hours (N=240) and our default architecture, we used a batch size of 128, the largest power of two that could fit on the GPU. We employed early-stopping to terminate training once there had been no improvement in the validation loss for three consecutive epochs, restoring the model checkpoint that achieved the minimum value. The training run which produced our best model took $6.4$ h. All models were implemented using the PyTorch~\citeSupp{paszke_pytorch_2019} framework.

\section{Transformer Output Tiling}\label{section:tiling}
Because of the quadratic complexity of the original Transformer~\citeSupp{vaswani_attention_2017}, we used an input sequence length of N=240 epochs, i.e. two hours, to the SleepVST model. This is much shorter than the sleep recordings within each dataset, which typically last between 8 and 12 hours. To apply SleepVST to these longer sequences, we re-applied the model at up to 30-minute steps, resulting in multiple outputs for timesteps away from the start and end of the recording. Using this approach, the model can be applied to 10 hours of waveform data in  $\approx 0.8$ s. 

\begin{figure}[htb]
  \centering
   \includegraphics[width=1\linewidth]{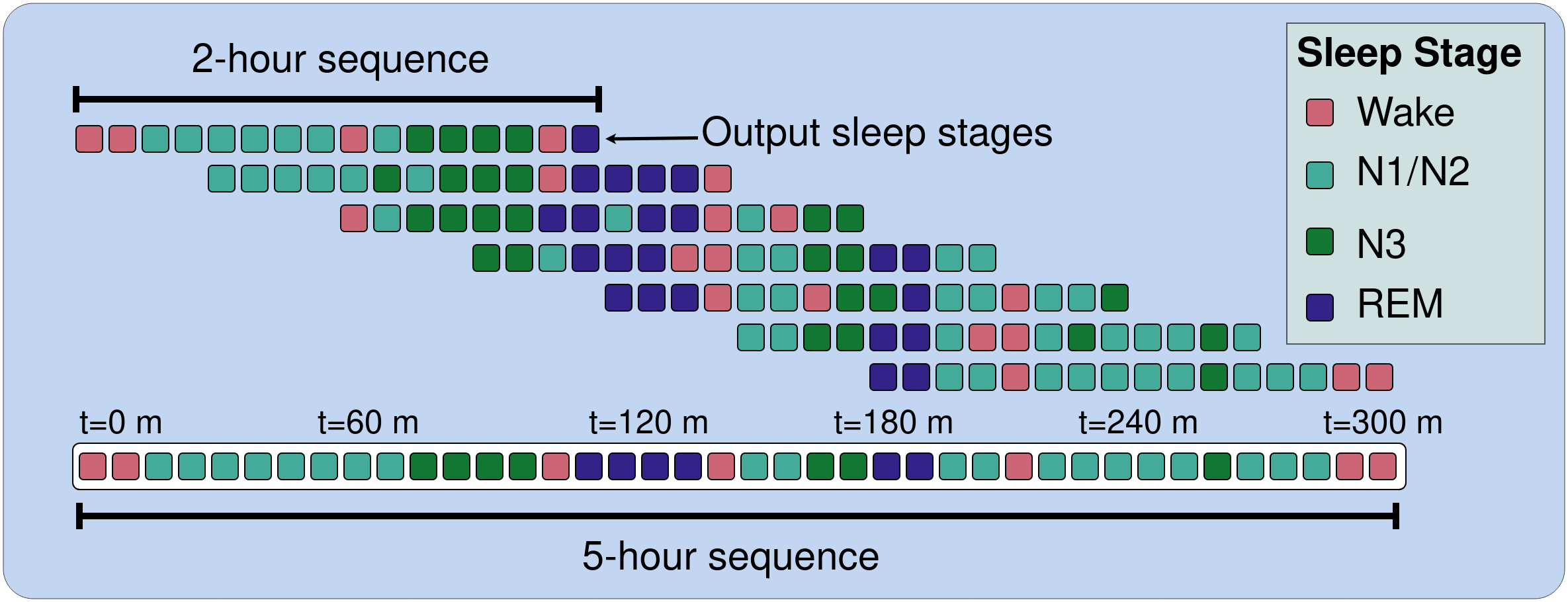}
   \caption{\textbf{Classifying longer input sequences using SleepVST.} After applying the model to overlapping two-hour sub-sequences, the modal classification at each timestep is used as the output classification.}\label{fig:tiling}
\end{figure}

When directly applying the pre-trained SleepVST model, e.g. to SHHS and MESA test sets, we took the mode of the overlapping classifications as the output classification. This is illustrated in \Cref{fig:tiling}.

\begin{figure}[htb]
  \centering
   \includegraphics[width=1\linewidth]{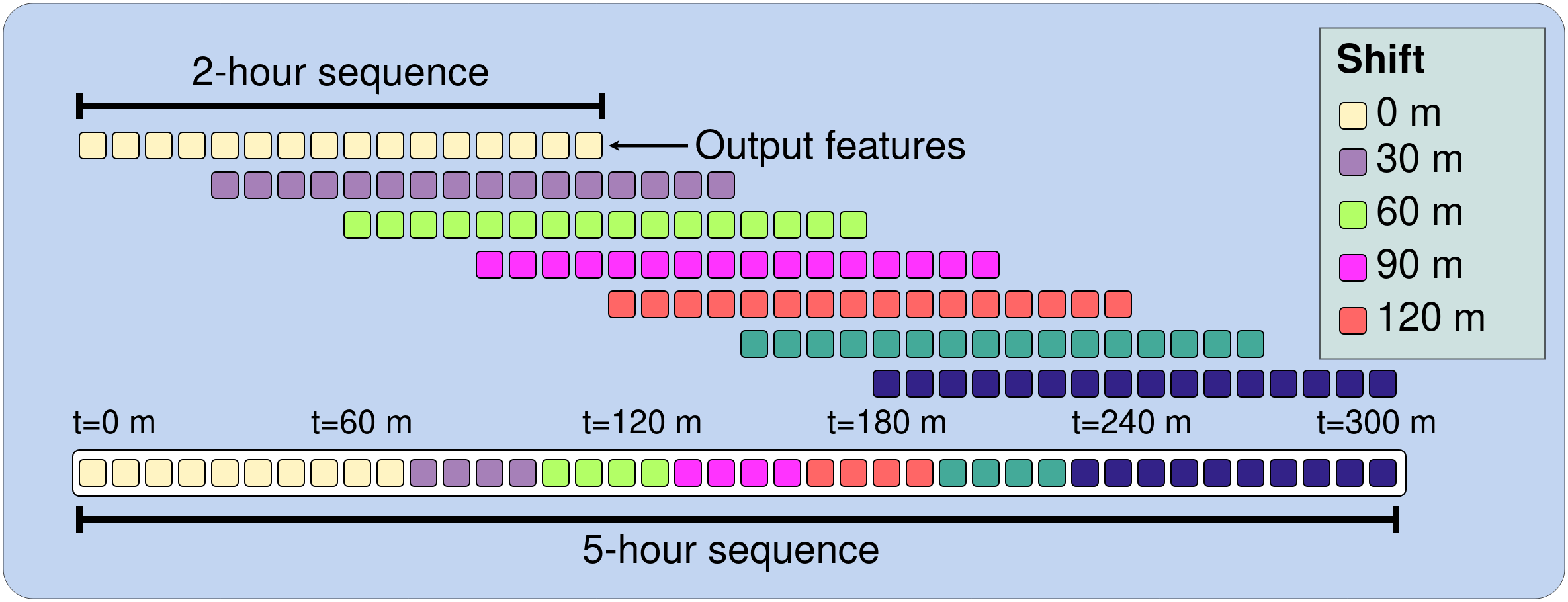}
   \caption{\textbf{Extracting features from longer input sequences using SleepVST.} The model is applied to two-hour input sub-sequences, producing overlapping output feature sequences. For each timestep, we use the feature vector which is closest to the middle of a sequence.}\label{fig:featuretiling}
\end{figure}

\begin{figure*}[htb]
  \centering
  \begin{subfigure}[b]{1\textwidth}
        \centering
        \includegraphics[width=0.9\linewidth]{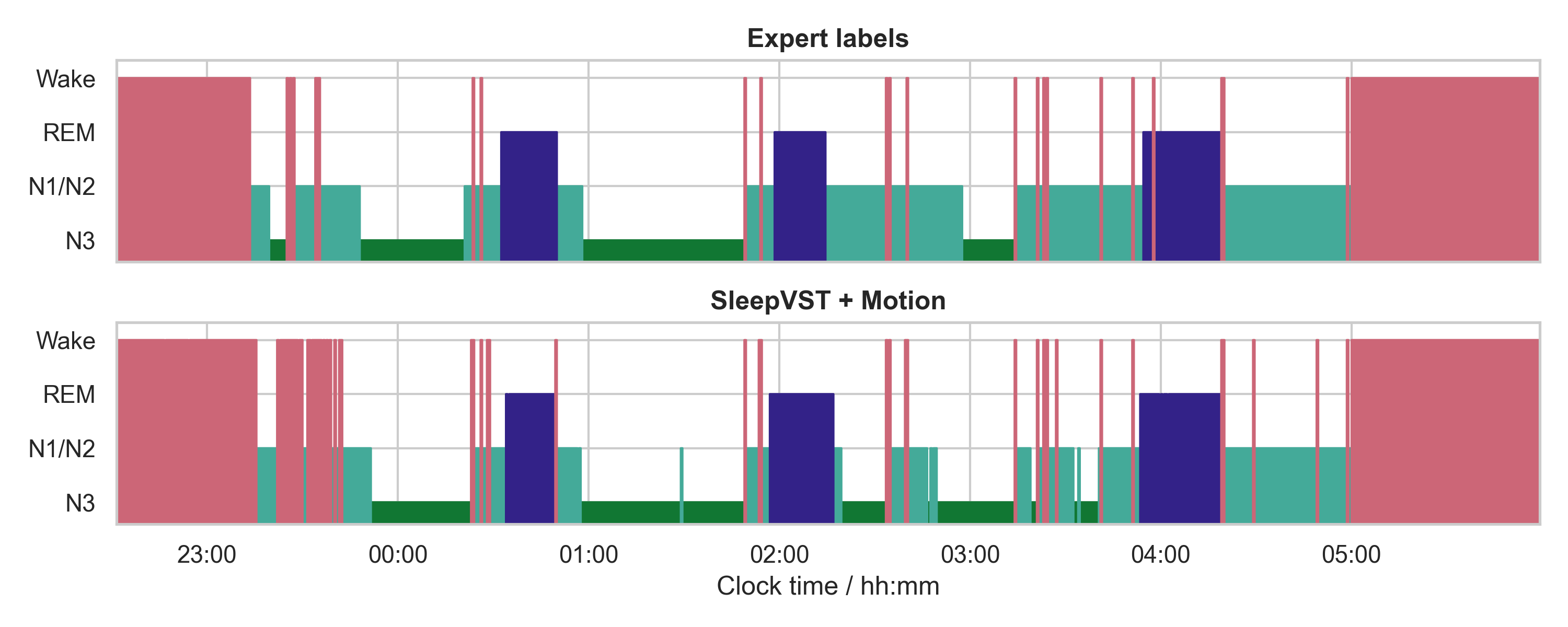}
        \caption{Cohen's $\kappa=0.84$ between model and expert labels.}
    \end{subfigure}\\
  \begin{subfigure}[b]{1\textwidth}
        \centering
        \includegraphics[width=0.9\linewidth]{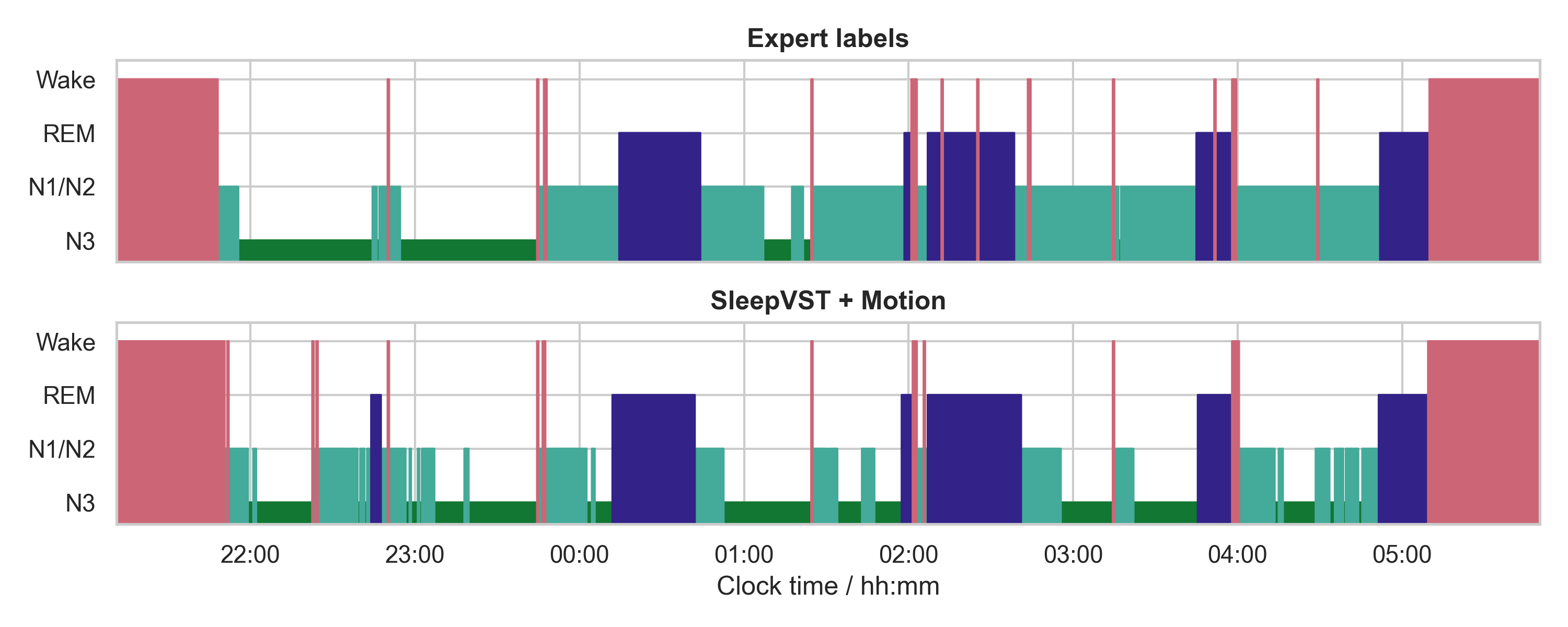}
        \caption{Cohen's $\kappa=0.64$ between model and expert labels.}
    \end{subfigure}\\
  \begin{subfigure}[b]{1\textwidth}
        \centering
        \includegraphics[width=0.9\linewidth]{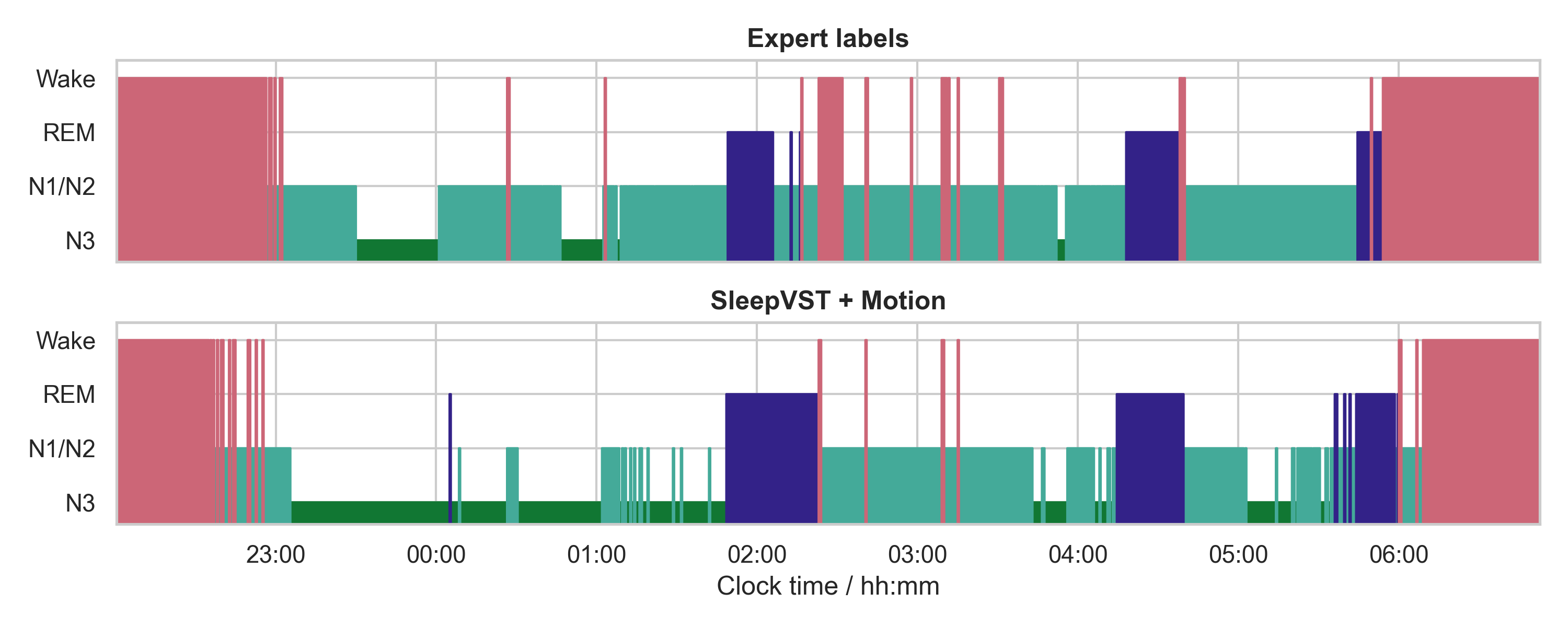}
        \caption{Cohen's $\kappa=0.39$ between model and expert labels.}
    \end{subfigure}%
   \caption{Example four-class sleep hypnograms from the OSV dataset for various model--expert Cohen's $\kappa$ agreement values. Within each subfigure, the top hypnogram shows expert labels annotated using signals from the vPSG recording, and the bottom hypnogram shows labels automatically generated from near-infrared video using our method.}\label{fig:extrahyps}
\end{figure*}

Similarly, when using SleepVST as a feature extractor on video-derived waveforms, we re-applied the model at intervals to produce overlapping two-hour sequences of feature vectors. For timesteps with multiple feature vectors, we used the feature vector which was closest to the middle of a sequence. This is illustrated in \Cref{fig:featuretiling}.

\section{Optical Flow Estimation}
We used the Dense Inverse Search algorithm~\citeSupp{kroeger_fast_2016} to calculate the optical flow field at a frequency of 4 Hz from the homography-transformed frames, following the same procedure as~\citeSupp{carter_deep_2023}. 

\section{Cohen's Kappa Calculation}
\label{section:kappa_definition}
From a confusion matrix $M\in\mathbb{R}^{C\times C}$ with elements $m_{ij}$, the function $K: \mathbb{R}^{C\times C}\rightarrow \mathbb{R}$ which calculates Cohen's $\kappa$ statistic is given by:
\begin{equation}
    K(M) = 1 - \frac{\sum_{i,j}w_{ij}m_{ij}}{\sum_{i,j}w_{ij}e_{ij}}
\end{equation}
where $w_{ij}$ and $e_{ij}$ are defined as follows:
\begin{align}
    w_{ij} &=  \begin{cases}
        0 & \text{if } i = j\\
        1 & \text{otherwise}
    \end{cases}\\
    e_{ij} &= \frac{\left(\sum_{i'}m_{i'j}\right)\cdot\left(\sum_{j'}m_{ij'}\right)}{\sum_{i'j'}m_{i'j'}}
\end{align}

\newpage
\section{SHHS Test Set}
\label{section:shhs_ids}
The 500 randomly sampled participant IDs used to form our test set are as follows:
\newline\newline
\noindent
{\tiny
203949, 204944, 200702, 203956, 202106, 201917, 203231, 204150, 205275, 203034, 201936, 205025, 200885, 201204, 204594, 201308, 204960, 205608, 204379, 203354, 204125, 204330, 203384, 201213, 201598, 204495, 204590, 203944, 203945, 202921, 201287, 203495, 205462, 204068, 203423, 202981, 203505, 204079, 204939, 203390, 204179, 204885, 202435, 202157, 202834, 200825, 203684, 205299, 200897, 200152, 202828, 203530, 203312, 205398, 203984, 202801, 203264, 201453, 203652, 200460, 202820, 204638, 203367, 205565, 200839, 203213, 204016, 204473, 200751, 201206, 200927, 201608, 201102, 205494, 201399, 200730, 202948, 200293, 204865, 203520, 204795, 200953, 203125, 205340, 205009, 204825, 200752, 202794, 203165, 203306, 203490, 204041, 202383, 204656, 203772, 203829, 204517, 201557, 202650, 203308, 203564, 202210, 200991, 205663, 203559, 204303, 200513, 205664, 202152, 204928, 203974, 200851, 205169, 205645, 204256, 201783, 201414, 204540, 204661, 203894, 201373, 202496, 200718, 202227, 200243, 203826, 202946, 203446, 202123, 200604, 201058, 205704, 204798, 205346, 204823, 203748, 203824, 200680, 200698, 203166, 204140, 205072, 203512, 200646, 205744, 200679, 204295, 201670, 204486, 203316, 203511, 200998, 204335, 200886, 202968, 204409, 205575, 202943, 205082, 203200, 200321, 205226, 200899, 203845, 202940, 200687, 200948, 203311, 200769, 200782, 203925, 200666, 202480, 201521, 205593, 200507, 203282, 204001, 205601, 203372, 204988, 204269, 203946, 205257, 200176, 204231, 204530, 204963, 200145, 200888, 202187, 200088, 203882, 203286, 204418, 200578, 204273, 200823, 203065, 205349, 203106, 200217, 202912, 203566, 200154, 201323, 200662, 204289, 205651, 203252, 205044, 201024, 203716, 204232, 200632, 200102, 205450, 200566, 202521, 202563, 200584, 203192, 201298, 205485, 205739, 204617, 204642, 201331, 205596, 203202, 203381, 204956, 203522, 200303, 203534, 204190, 201130, 201268, 200191, 201513, 200955, 203689, 203157, 201401, 201517, 202489, 203018, 203232, 205146, 202963, 202821, 204287, 204422, 204472, 200334, 200925, 203135, 200516, 202442, 204171, 203392, 201503, 202458, 205587, 203502, 203626, 204702, 204599, 200150, 205605, 202566, 204296, 202221, 203296, 205537, 203860, 200842, 200318, 204023, 202463, 201628, 200858, 205419, 201068, 205312, 202663, 202444, 200579, 201629, 203198, 204283, 204846, 204699, 200178, 200981, 203237, 200564, 204340, 202785, 202938, 204506, 201493, 203610, 201353, 203769, 200105, 204522, 204343, 204086, 200320, 204425, 202842, 204459, 202226, 203528, 204978, 204461, 204691, 200111, 201083, 200950, 200108, 203455, 204647, 200952, 204443, 204435, 204504, 205064, 203476, 203695, 203721, 202201, 200841, 200935, 204093, 201223, 201470, 204747, 205350, 204871, 203303, 204132, 201219, 204676, 201402, 205722, 204115, 200744, 205772, 203281, 204926, 205086, 202990, 203235, 204384, 201432, 203039, 200387, 203961, 203456, 203462, 202546, 203895, 205595, 204496, 201241, 205004, 200712, 204565, 203671, 200936, 201271, 203734, 202428, 203260, 200653, 204405, 201982, 200703, 203314, 202405, 200668, 205702, 204431, 202608, 203056, 204544, 203754, 205761, 203942, 200835, 200192, 205539, 200945, 203254, 204177, 200219, 205486, 202642, 202417, 203498, 203347, 204759, 202361, 205255, 201064, 201544, 200295, 204233, 204187, 202150, 204224, 204370, 204952, 200571, 203976, 204304, 200347, 205126, 203224, 200591, 203060, 201543, 204923, 201778, 204914, 205222, 204907, 200766, 205305, 204235, 200406, 205661, 200209, 204778, 201640, 204236, 200901, 205356, 200853, 200210, 204898, 203269, 203451, 202825, 201299, 204690, 203557, 203589, 203037, 204337, 204460, 201316, 201312, 204856, 203138, 203412, 200242, 203460, 200920, 200887, 201918, 203395, 205530, 201349, 200829, 203208, 201519, 200386, 203117, 200466, 202605, 203121, 200624, 205721, 204323, 204554, 205289, 204934, 200233, 203149, 204170, 203966, 205252, 205548, 203006, 202902, 203818, 202942, 201018, 205588, 202395, 203709, 205591, 205532, 204763, 202829, 205626, 201538.
}
\end{document}